\documentclass[11pt, a4paper, logo, copyright, nonumbering]{mll_style}

\usepackage{amsmath,amsfonts,bm}

\def\eqref#1{equation~\ref{#1}}

\def\1{\bm{1}}

\DeclareMathAlphabet{\mathsfit}{\encodingdefault}{\sfdefault}{m}{sl}
\SetMathAlphabet{\mathsfit}{bold}{\encodingdefault}{\sfdefault}{bx}{n}

\DeclareMathOperator*{\argmin}{arg\,min}

\PassOptionsToPackage{round}{natbib}
\usepackage{natbib}
\usepackage[most]{tcolorbox}
\usepackage{nicefrac}
\usepackage{multirow}
\usepackage{subcaption}
\usepackage{wrapfig}
\usepackage{enumitem}
\usepackage{algorithm}
\usepackage{algpseudocode}

\usepackage[toc,page,header]{appendix}
\usepackage{minitoc}

\renewcommand \partname{}

\newcommand{\viewsuite}{\textsc{ViewSuite}}
\newcommand{\ie}{\textit{i.e.}}

\title{\centering
\vspace{-10pt}
Planning with the Views 
}

\date{}

\author[*]{
\vspace{-10pt}
{Kangrui Wang$^{1}$, Linjie Li$^{2}$, Zhengyuan Yang$^{3}$, Shiqi Chen$^{4}$, Zihan Wang$^{1}$,  Li Fei-Fei$^{5}$, \newline Jiajun Wu$^{5}$, Leonidas Guibas$^{5}$, Lijuan Wang$^{3}$, Manling Li$^{1}$}
\\
\small $^1$Northwestern University~~~
$^2$University of Washington~~~
$^3$Microsoft\\
\small $^4$University of Oxford~~~
$^5$Stanford University
\vspace{-20pt}
}

\begin{abstract}
Can VLMs predict how each camera move changes the view, and plan many such moves ahead? 
We call this ability \emph{view planning}: using camera moves as planning primitives to find a target view in 3D.
We study view planning in \viewsuite{}, built on ScanNet~\citep{scannet}
with full 6-DoF camera pose control, and decompose it into two abilities: \emph{tracking} how given camera actions change the
view, and \emph{composing} a path that localizes an unseen target view. Across 13 frontier VLMs, a sharp planning gap emerges: models track local view transitions but collapse when they must plan toward
an unseen target view. 
This planning inability cannot simply be fixed by reinforcement learning (RL):
with success near $2.5\%$, reward is too sparse for RL to bootstrap. 
Our key insight is to distill valid view transitions from on-policy self-exploration trajectories, by aggregating them into a \emph{view graph} and distilling it into supervised demonstrations. 
Without any stronger teacher, this lifts Qwen2.5-VL-7B from $2.5\%$ to $47.8\%$ on interactive
view planning, surpassing GPT-5.4 Pro ($19.9\%$) and Gemini 3.1 Pro ($21.3\%$).
View planning offers a clean probe for prospective spatial reasoning: mentally looking ahead, predicting how future viewpoint changes will reshape observation, and inferring the camera pose of a target view before it is fully observed. Frontier VLMs still lack this capability, 
and the learned spatial priors transfer to other view-understanding tasks. 
\par\vspace{8pt}
\noindent

{\footnotesize\normalfont
\setlength{\tabcolsep}{0pt}%
\tabular{@{}l@{\hspace{0.55em}}l@{\hspace{1.4em}}l@{}}
\textbf{Website:} \href{https://viewagent.github.io/}{\nolinkurl{https://viewagent.github.io}} \\
 \textbf{Code:} \href{https://github.com/mll-lab-nu/viewagent}{\nolinkurl{https://github.com/mll-lab-nu/viewagent}} \\
\textbf{Models:}  \href{https://huggingface.co/collections/MLL-Lab/viewagent-models}{\nolinkurl{https://huggingface.co/collections/MLL-Lab/viewagent-models}} \\
\textbf{Data:} \href{https://huggingface.co/collections/MLL-Lab/viewagent-datasets}{\nolinkurl{https://huggingface.co/collections/MLL-Lab/viewagent-datasets}} \\
\endtabular
}
\end{abstract}

\begin{document}

\doparttoc
\faketableofcontents

\maketitle

\section{Introduction}
\label{sec:introduction}


To understand a 3D scene, an agent must be able to situate a view within it: infer where the camera stands, actively choose where to look next, and reason about how moving the camera can bring another view into sight. We call this \textbf{View Planning}, localizing a target view by composing a path of camera actions. 
It factors into two coupled abilities: \emph{tracking} how a path changes the view, and actively \emph{composing} a path to find where an unseen target view was taken. Intuitively, this is the difference between reading a given path and composing one from scratch. The first is local, reasoning relative to the current view. The second is global, requiring \emph{prospective spatial reasoning}: anchoring the current and target views in a shared allocentric frame, mentally rolling out how camera actions update observations, and looking ahead through a sequence of view transitions to compose a multi-turn plan toward the target view.

It is an instance of active perception~\citep{bajcsy1988active, aloimonos1988active, bajcsy2018revisiting}, where an agent actively chooses where to look and how to move the camera to change its view. 
Prior view search has advanced from 2D image regions~\citep{vstar,actiview} to $360^{\circ}$ panoramas~\citep{thinking360}. We take the next step: \textit{multi-step view planning in real 3D scenes}, where the agent controls the full 6-DoF camera pose. Success is scored by how accurately the agent estimates the target view's camera pose, not by whether it physically reaches an object or a physical destination. Thus, view planning targets a capability beyond navigation: mentally simulating an unobserved viewpoint and localizing it only by what it shows. The core challenge is not simply to move through views, but to plan with views: to mentally look ahead, forecast how actions change future observations, and infer where the target view lies before it is fully observed. The agent must use views as \emph{planning primitives}.

Our setting differs from previous view search in three ways (Table~\ref{tab:benchmark_comparison}): (1) \emph{real 3D scenes} rather than synthetic graphics; (2) \emph{6-DoF camera pose control}, where the camera moves with full six degrees of
freedom, different from physical affordance, embodied navigation, and 2D image cropping; and (3) \emph{multi-turn view composition} rather than single-step decisions.

\begin{table}[t]
\centering
\caption{Comparison with existing view reasoning benchmarks. \viewsuite{} takes the next step: multi-turn view planning in real-world 3D scenes with full 6-DoF camera pose control, at $165$K instances. $^{*}$EmbodiedBench supports multiple action types. Dashes: not applicable or not reported.}
\label{tab:benchmark_comparison}
\small
\setlength{\tabcolsep}{2pt}
\begin{tabular}{@{}lllllll@{}}
\toprule
Benchmark & Task & Scale & Real World & 3D & Action Space & Multi-turn \\
\midrule
ViewSpatial-Bench~\citep{viewspatialbench} & View-Centric QA & 5.7K & \checkmark & & -- & \\
VSI-Bench~\citep{vsibench} & Video QA & 5K & \checkmark & \checkmark & -- & \\
CameraBench~\citep{camerabench} & Video QA & 3K & \checkmark & & -- & \\
MindCube~\citep{mindcube} & View-Centric QA & 21K & & \checkmark & -- & \\
V$^\star$~\citep{vstar} & Visual Search & -- & \checkmark & & 2-DoF & \checkmark \\
ActiView~\citep{actiview} & Visual Search & 3K & \checkmark & & 2-DoF & \checkmark \\
H$^\star$Bench~\citep{thinking360} & Visual Search & -- & \checkmark & \checkmark & 2-DoF & \checkmark \\
HM3D-OVON~\citep{hm3dovon} & Embodied Agent & -- & \checkmark & \checkmark & 4-DoF & \checkmark \\
EmbodiedBench~\citep{embodiedbench} & Embodied Agent & 1.1K & & \checkmark & 6-DoF$^{*}$ & \checkmark \\
Theory of Space~\citep{theoryofspace} & Embodied QA & 2.7K & & \checkmark & 2-DoF & \checkmark \\
\textbf{\viewsuite{} (Ours)} & Visual Search & 165K & \checkmark & \checkmark & 6-DoF & \checkmark \\
\bottomrule
\end{tabular}
\end{table}

\begin{figure*}[t]
\centering
\includegraphics[width=\textwidth]{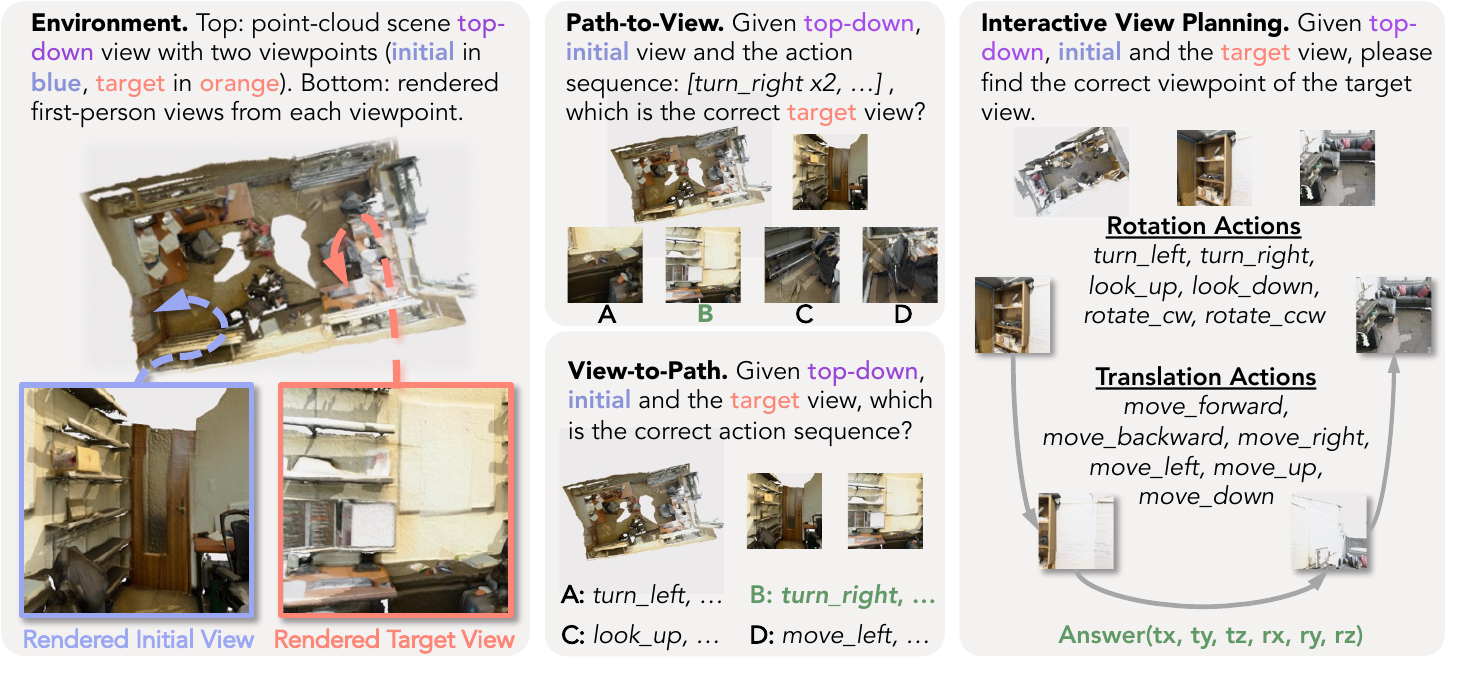}
\caption{Overview of \viewsuite{}. \textbf{Left}: Point cloud environment built on ScanNet with rendered first-person views. \textbf{Middle}: Path-to-View (P2V; predict the resulting view from an action sequence) and View-to-Path (V2P; infer the action sequence from two views), both single-turn. \textbf{Right}: Interactive View Planning (IVP), a multi-turn task where the agent plans view changes to localize the target view and submits a camera pose estimate.}
\label{fig:environment_overview}
\vspace{-7pt}
\end{figure*}

We study whether current VLMs have this ability by building \viewsuite{}, a view-planning
environment with full 6-DoF camera pose control on $286$ real ScanNet~\citep{scannet} indoor scenes, yielding ${\sim}55$K view pairs and ${\sim}165$K task instances across three diagnostic tasks\footnote{The IVP success threshold is calibrated against human judgments via an alignment study (Appendix~\ref{app:threshold_calibration}).}. As shown in Figure~\ref{fig:environment_overview}, 
in \textbf{\underline{Path-to-View}} (P2V) and \textbf{\underline{View-to-Path}} (V2P), the path is given, explicitly in the question or among the answer
choices. In a \textit{single turn}, VLMs need only \emph{read} it: predict the view an action sequence
produces, or infer which sequence connects two views. 
In \textbf{\underline{Interactive View Planning}} (IVP), 
no path is given: VLMs must \emph{compose} one, issue actions over \textit{multiple turns}
to localize an unseen target view, then submit a 6-DoF estimate of it. 
P2V and V2P thus ask whether VLMs can understand view transitions, while IVP asks whether they can use that understanding, an any-view-to-any-view problem that subsumes the tracking the first two require and adds the harder demand of placing a target view they have not yet seen.

Evaluating $13$ frontier VLMs reveals a sharp planning gap: the best models can track local view changes over a short horizon ($\sim\!70\%$ on P2V/V2P) but collapse when they must compose them into a plan (at most $21.3\%$ and most below $10\%$ on IVP). 
The failure is driven neither by exploration budget nor by rendering fidelity: more turns do not close the gap, and a higher-fidelity renderer leaves IVP essentially unchanged. What predicts failure is view distance: rotation distance for the tracking tasks, position distance for planning. 
The few successes reveal a sharper pattern: across models, over $90\%$ of successful IVP runs occur only after the target view becomes observable. In other words, models usually solve IVP by moving until they happen to see the target view, then matching it, rather than by inferring where the target view lies beforehand. The planning gap is therefore more precisely a \emph{cognitive gap}: even with the global top-down map in hand, frontier VLMs can rarely anchor egocentric views onto the map, mentally simulate how camera actions change those views, or localize a target view before seeing it. Prospective spatial reasoning is thus a harder, higher-level capability than tracking.
 
A natural way to learn planning is reinforcement learning (RL), yet this gap makes RL surprisingly ineffective. 
With a naive policy succeeding only ${\sim}2.5\%$, reward is too sparse to reinforce: direct PPO plateaus at $3.2\%$, GRPO with reward-variance filtering reaches
$5.2\%$, and bootstrapping from successful trajectories (SFT) reaches $6.2\%$. 
Our way past this bottleneck comes from a simple observation: every trajectory, successful or not, traces valid view transitions. 

Distilling valid view transitions from raw exploration is then the central challenge. We construct a \emph{view graph}, an any-view-to-any-view map assembled from the agent's on-policy self-exploration. We then distill it into supervised demonstrations for view planning, and alternate this distillation with further self-exploration. As the policy improves, its exploration grows the view graph outward iteration by iteration, and the resulting supervision remains matched to the region of view space the agent can actually plan over. 
Like on-policy distillation~\citep{lu2025onpolicydistillation}, our framework learns from the agent's on-policy exploration; unlike it, there is no stronger teacher to imitate, and the teacher is the environment itself, whose structure the agent reveals by moving through it.

This iterative self-exploration and view-graph distillation lifts Qwen2.5-VL-7B from $2.5\%$ to
$47.8\%$ on IVP, surpassing GPT-5.4 Pro ($19.9\%$) and Gemini 3.1 Pro ($21.3\%$).
Building the view graph from random rather than on-policy trajectories reaches only $13.0\%$,
confirming that effective supervision must grow with the policy. 
Beyond these, we show that interactive view planning yields \emph{transferable spatial priors}: under identical post-training, our model outperforms its base counterpart on other related view-understanding tasks. 
Together, view planning moves toward prospective spatial cognition~\citep{tolman1948,schacter2007}: whether VLMs can internalize a world model of view transitions, plan over possible future views, and localize an unseen viewpoint.  



\section{\viewsuite{}: Problem Formulation, Environment, and Benchmark}
\label{sec:viewsuite}

\viewsuite{} casts view planning as a multi-turn decision process: an agent issues camera actions that alter its 6-DoF pose in a 3D point-cloud environment built on real ScanNet~\citep{scannet} indoor scenes, and at termination submits a camera-pose estimate of the target view, scored by distance to the ground truth (formal MDP and reward in Section~\ref{sec:ivet}, Eq.~\ref{eq:reward}).

\subsection{Problem Formulation}
\label{sec:problem_formulation}

We decompose view planning into two coupled abilities: (1)~\emph{tracking} how a given path changes the view, and (2)~\emph{composing} a path that localizes an unseen target view. We design three complementary tasks targeting these abilities: P2V and V2P test tracking when the path is given, while IVP requires composing a plan when it is not.

\vspace{-7pt}
\paragraph{Path-to-View (P2V).}
Given an initial view, a top-down reference, and an action sequence, the model predicts the resulting view from four options (multiple-choice), testing whether it can mentally simulate view transitions (Figure~\ref{fig:p2v_example}).

\vspace{-7pt}
\paragraph{View-to-Path (V2P).}
Given initial and target views plus a top-down view, the model identifies which action sequence was executed, again from four options. P2V and V2P together test view-transition tracking in both directions (Figure~\ref{fig:v2p_example}).

\vspace{-7pt}
\paragraph{Interactive View Planning (IVP).}
Given an initial view, a target view, and a top-down reference, the agent issues multiple actions per turn, observes the resulting view and camera pose, and within a fixed turn budget submits a 6-DoF camera pose estimate of where the target view was taken. Unlike the single-turn P2V and V2P, IVP requires the agent to plan a sequence of view changes over multiple turns to localize the target view (Figure~\ref{fig:tvn_example}).

\subsection{Camera Pose Control Interface}
\label{sec:environment}

{\sloppy
\viewsuite{} exposes 6-DoF camera pose control through $12$ step-size-parameterized actions: six translations (\texttt{move\_forward}, \texttt{move\_backward}, \texttt{move\_left}, \texttt{move\_right}, \texttt{move\_up}, \texttt{move\_down}) and six rotations (\texttt{turn\_left}, \texttt{turn\_right}, \texttt{look\_up}, \texttt{look\_down}, \texttt{rotate\_cw}, \texttt{rotate\_ccw}); full geometric definitions are in Appendix~\ref{app:action_space} (Figure~\ref{fig:environment_overview}).\par} Step sizes are discretized so that a model controls the camera by \emph{selecting} actions rather than specifying precise motion parameters, while each action still produces a visibly distinct view. The interface is decoupled from the 3D backend (point clouds, meshes, or simulators) and renders with Open3D~\citep{open3d}.

\subsection{Data Collection and Evaluation}
\label{sec:data}
To construct the task data, we sample initial-target view pairs from ScanNet video frames.
We use fixed step sizes $s_t {=} 0.5\,\text{m}$ for translation and $s_r {=} 30^\circ$ for rotation.
We conduct scene-level and pair-level filtering, then reformat each pair into P2V, V2P, and IVP instances (detailed in Appendix~\ref{app:data_pipeline}). This yields $\sim\!55$K view pairs across $286$ ScanNet scenes.

\vspace{-7pt}
\paragraph{Dataset splits.}
From the $\sim\!55$K view pairs, we split pairs within each scene at $1{:}10$ ratio to form \viewsuite{}-5K and \viewsuite{}-50K; all experiments in this paper use \viewsuite{}-5K, with the full set reserved for future scaling studies. Scenes are then partitioned into train/dev/test at $8{:}1{:}1$ ratio, ensuring no scene overlap across splits. Each view pair yields three task instances (P2V, V2P, and IVP), giving $\sim\!15$K instances in \viewsuite{}-5K, with the test set containing $530$ pairs $\times\,3$ tasks $= 1{,}590$ instances. In total, \viewsuite{} provides $\sim\!165$K task instances.

\vspace{-7pt}
\paragraph{Evaluation metrics.}
For P2V and V2P, we use accuracy (correct option selected). For IVP, we use \emph{Success Rate}, defined by pose distance. We extract 6-DoF camera poses from camera-to-world extrinsic matrices, decomposing each into a position vector $\mathbf{t} \in \mathbb{R}^3$ and a rotation matrix $R \in \mathrm{SO}(3)$.
Translation distance and rotation distance between two poses are
\begin{equation}
  d_\text{pos} = \| \mathbf{t}_1 - \mathbf{t}_2 \|_2, \quad
  d_\text{rot} = \arccos\!\Bigl(\tfrac{\mathrm{tr}(R_1^\top R_2)-1}{2}\Bigr).
\end{equation}
The agent succeeds when $d_\text{pos} \leq \beta_t \cdot s_t$ and $d_\text{rot} \leq \beta_r \cdot s_r$ for threshold multipliers $\beta_t, \beta_r$.
We calibrate $\beta_t$ and $\beta_r$ against human judgments of whether two rendered views depict the same place, selecting the combination that maximizes $F_1$ agreement with those judgments (full study in Appendix~\ref{app:threshold_calibration}).
The best setting is $\beta_t {=} 1,\, \beta_r {=} 1$, i.e., one step size in each dimension (details in Table~\ref{tab:threshold_calibration}).

\vspace{-7pt}
\paragraph{Unified view distance.}
We combine translation and rotation distance into a single difficulty score, the \emph{unified view distance}:
\begin{equation}
  d = \sqrt{(d_\text{pos}/s_t)^2 + (d_\text{rot}/s_r)^2},
\end{equation}
where normalizing by the step sizes makes each unit of $d$ approximately correspond to one atomic action, so $d$ approximates the length of the shortest action plan connecting two views.
Across the $530$ test pairs, $d$ ranges from $1.4$ to $6.8$ with mean $3.7$ (Figure~\ref{fig:view_distance_dist} in Appendix~\ref{app:viewsuite}).
We split test pairs into \textsc{Short} ($d < 3$; $185$ pairs) and \textsc{Long} ($d \geq 3$; $345$ pairs) subsets, and further decompose along the translation and rotation axes for finer-grained diagnosis in Section~\ref{sec:eval_analysis}.

\section{Planning Gap in Frontier VLMs}
\label{sec:evaluation}

\subsection{Single-turn Tracking Understood, Multi-turn Planning Collapses}
\label{sec:eval_main}

We evaluate $13$ frontier VLMs ($7$ proprietary, $6$ open-weight) plus a random-response baseline, detailed in Table~\ref{tab:main_results}.
The central finding is a planning gap: frontier VLMs track local view transitions but cannot compose them into a multi-turn plan that localizes a target view.
On P2V and V2P, the best models achieve a modest $\sim\!50\%$ overall, rising to over $70\%$ on short-horizon samples, well above the $25\%$ MCQ chance baseline but far from solving the task. This indicates non-trivial but partial knowledge of view-action mappings, which degrades further on long-horizon samples that require mentally simulating cumulative transformations.
When the model must instead compose the path itself, performance collapses. On IVP the best model (Gemini 3.1 Pro) reaches only $21.3\%$, most models fall below $10\%$, and on long-horizon samples most fall below $3\%$; the gap holds across proprietary and open-weight models, with every open-weight model below $5\%$.
Notably, GPT-5.4 Pro outperforms GPT-5.4 across all tasks and splits, including IVP; given that GPT-5.4 Pro is widely believed to be a test-time scaled variant of GPT-5.4, this suggests that additional test-time computation can meaningfully improve spatial reasoning on \viewsuite{}.

\begin{table}[t]
\centering
\caption[Main evaluation results on \viewsuite{}.]{\textbf{The planning gap in frontier VLMs.} Main evaluation results on \viewsuite{}: accuracy (\%) for P2V/V2P and success rate (\%) for IVP\protect\footnotemark{}, on Short (view distance $d{<}3$) / Long ($d{\ge}3$) / Overall splits. The best models exceed $70\%$ on short-horizon P2V/V2P but reach at most $21.3\%$ on IVP, with most below $10\%$. Proprietary: GPT~\citep{gpt5,gpt54}, Gemini~\citep{gemini3,gemini31}, Claude~\citep{claude46}, Grok~\citep{grok420}. Open-weight: Qwen~\citep{qwen25vl,qwen3vl,qwen35}, GLM~\citep{glmv}, Kimi~\citep{kimik25}. Models sorted by Overall within each group; best per column in \textbf{bold}.}
\label{tab:main_results}
\small
\setlength{\tabcolsep}{3.5pt}
\begin{tabular}{@{}l ccc ccc ccc c@{}}
\toprule
& \multicolumn{3}{c}{Path-to-View (P2V)} & \multicolumn{3}{c}{View-to-Path (V2P)} & \multicolumn{3}{c}{Interactive View Planning (IVP)} & \\
\cmidrule(lr){2-4} \cmidrule(lr){5-7} \cmidrule(lr){8-10}
Model & Short & Long & All & Short & Long & All & Short & Long & All & Overall \\
\midrule
\rowcolor{gray!15}
Random Response & 20.7 & 24.6 & 23.3 & 24.3 & 26.5 & 25.7 & 2.2 & 0.0 & 0.8 & 16.6 \\
\midrule
\multicolumn{11}{l}{\textit{Proprietary Models}} \\
\midrule
GPT-5.4 Pro & \textbf{70.8} & \textbf{43.8} & \textbf{53.2} & \textbf{72.4} & 38.8 & \textbf{50.6} & \textbf{34.8} & 11.7 & 19.9 & \textbf{41.2} \\
Gemini 3.1 Pro & 63.8 & 40.9 & 48.9 & 53.0 & \textbf{47.5} & 49.4 & 28.6 & \textbf{17.4} & \textbf{21.3} & 39.9 \\
GPT-5.4 & 57.3 & 42.9 & 47.9 & 60.5 & 37.4 & 45.5 & 33.5 & 7.5 & 16.6 & 36.7 \\
Grok 4.20 Beta & 61.6 & 38.0 & 46.2 & 44.9 & 44.3 & 44.5 & 17.3 & 2.9 & 7.9 & 32.9 \\
GPT-5.1 & 60.5 & 35.1 & 44.0 & 52.4 & 33.3 & 40.0 & 11.9 & 3.2 & 6.2 & 30.1 \\
Claude Opus 4.6 & 46.5 & 28.4 & 34.7 & 47.6 & 38.3 & 41.5 & 23.8 & 3.8 & 10.8 & 29.0 \\
Gemini 3 Pro & 50.3 & 31.0 & 37.7 & 44.9 & 35.4 & 38.7 & 13.5 & 7.0 & 9.2 & 28.5 \\
\midrule
\multicolumn{11}{l}{\textit{Open-Weight Models}} \\
\midrule
Qwen3.5-397B & \textbf{57.8} & 30.1 & \textbf{39.8} & \textbf{44.3} & \textbf{31.0} & \textbf{35.7} & \textbf{12.4} & 0.0 & \textbf{4.3} & \textbf{26.6} \\
GLM-4.6V & 36.4 & 23.2 & 27.8 & 31.4 & 29.7 & 30.2 & 9.2 & \textbf{1.2} & 4.0 & 20.7 \\
Qwen2.5-VL-72B & 28.1 & 29.3 & 28.9 & 35.7 & 30.1 & 32.1 & 2.2 & 0.6 & 1.1 & 20.7 \\
Qwen3-VL-32B & 27.0 & 27.5 & 27.4 & 41.1 & 28.7 & 33.0 & 4.3 & 0.0 & 1.5 & 20.6 \\
Kimi K2.5 & 36.2 & 24.6 & 28.7 & 18.4 & 29.3 & 25.5 & 4.9 & \textbf{1.2} & 2.5 & 18.9 \\
Qwen2.5-VL-7B & 23.8 & \textbf{32.5} & 29.4 & 27.0 & 22.9 & 24.3 & 7.0 & 0.0 & 2.5 & 18.7 \\
\bottomrule
\end{tabular}
\end{table}
\footnotetext{GPT-5.4 Pro refuses $23$ of the $530$ IVP test instances under its violation policy. Its IVP rates are reported over the remaining $507$ valid instances ($101/507{=}19.9\%$). All other models are evaluated on the full $530$.}

\subsection{What Bottlenecks Interactive View Planning?}
\label{sec:eval_analysis}

\paragraph{Does turn budget affect IVP performance?}
A natural hypothesis is that models fail at IVP simply because $10$ turns is insufficient.
We test this by increasing the turn budget to $20$ and $30$ for four proprietary models (Table~\ref{tab:budget_results}).
All models improve from $10$ to $20$ turns, with Claude Opus 4.6 showing the largest gain (nearly doubling).
However, gains from $20$ to $30$ turns are marginal or zero for most models.
This diminishing return suggests that IVP performance is bottlenecked by planning ability rather than exploration horizon, as models exhaust their effective strategies well before the turn limit.

\begin{table}[t]
\centering
\caption{\textbf{Neither more turns nor sharper rendering closes the planning gap.} Left three blocks: interactive view planning (IVP) success rate (\%) at turn budgets of $B{=}10$, $20$, and $30$ with point-cloud rendering; gains from $20$ to $30$ turns are marginal. Rightmost block: P2V / V2P / IVP All-split results at a turn budget of $10$ with higher-fidelity Gaussian Splatting (GS) rendering; IVP improves only marginally.}
\label{tab:budget_results}
\small
\setlength{\tabcolsep}{3pt}
\begin{tabular}{@{}l ccc ccc ccc cccc@{}}
\toprule
& \multicolumn{3}{c}{IVP, B = 10} & \multicolumn{3}{c}{IVP, B = 20} & \multicolumn{3}{c}{IVP, B = 30} & \multicolumn{4}{c}{GS, B = 10} \\
\cmidrule(lr){2-4} \cmidrule(lr){5-7} \cmidrule(lr){8-10} \cmidrule(lr){11-14}
Model & Short & Long & All & Short & Long & All & Short & Long & All & P2V & V2P & IVP & Overall \\
\midrule
Gemini 3.1 Pro & 28.6 & \textbf{17.4} & \textbf{21.3} & 30.8 & \textbf{18.8} & \textbf{23.0} & 33.0 & \textbf{17.9} & \textbf{23.2} & \textbf{55.3} & \textbf{49.4} & \textbf{23.2} & \textbf{42.6} \\
GPT-5.4 & \textbf{33.5} & 7.5 & 16.6 & 33.0 & 11.9 & 19.2 & 35.1 & 12.1 & 20.2 & 43.8 & 31.1 & 18.5 & 31.1 \\
Grok 4.20 Beta & 17.3 & 2.9 & 7.9 & 23.8 & 5.2 & 11.7 & 27.0 & 3.5 & 11.7 & 28.3 & 31.5 & 8.1 & 22.6 \\
Claude Opus 4.6 & 23.8 & 3.8 & 10.8 & \textbf{31.9} & 10.1 & 17.7 & \textbf{35.7} & 11.0 & 19.6 & 35.3 & 41.3 & 12.3 & 29.6 \\
\bottomrule
\end{tabular}
\end{table}

\vspace{-7pt}
\paragraph{Does rendering quality affect model performance?}
A natural concern is that point-cloud rendering, with its sparse and noisy pixels, may itself bottleneck the agent.
We re-render the test set with 3D Gaussian Splatting~\citep{gsplat}, a higher-fidelity neural renderer, using pretrained per-scene 3DGS reconstructions of the ScanNet scenes from SceneSplat-7K~\citep{scenesplat}, and re-evaluate four proprietary models on all three tasks at a turn budget of $10$ (Table~\ref{tab:budget_results}, rightmost block).
The pattern across tasks is asymmetric: IVP improves consistently but only marginally ($+0.2$ to $+1.9$ points), whereas P2V and V2P show mixed and sometimes large changes: Gemini 3.1 Pro gains $+6.4$ on P2V, while GPT-5.4 and Grok 4.20 Beta lose $14.4$ and $13.0$ on V2P (relative to their point-cloud rendering scores in Table~\ref{tab:main_results}).
That a higher-fidelity renderer does not unlock single-turn performance, and yields only modest gains on IVP, indicates that the IVP bottleneck is not the visual fidelity of the rendered observation but the model's ability to compose view changes into a multi-turn plan. Combined with the test-time-compute gain above (GPT-5.4 Pro over GPT-5.4), this places the bottleneck in reasoning rather than perception: sharper pixels do not help, more thinking does.

\begin{figure*}[t]
\centering
\includegraphics[width=\textwidth]{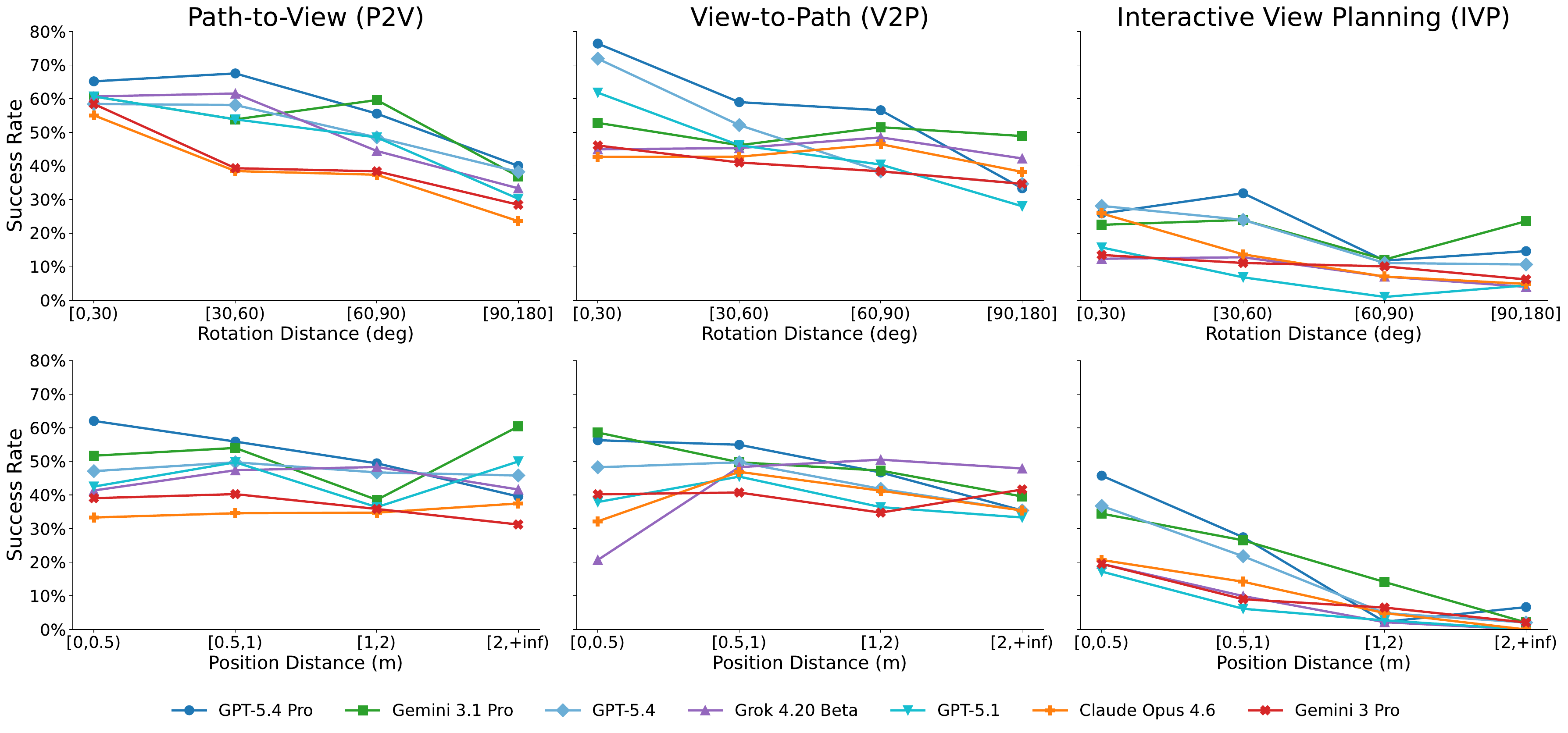}
\caption{\textbf{Tracking degrades with rotation distance, planning with position distance.} Success rate vs.\ rotation distance (top) and position distance (bottom) for proprietary models across all three tasks: P2V/V2P degrade primarily with rotation distance, while IVP success collapses with position distance.}
\label{fig:dual_axis}
\end{figure*}

\vspace{-8pt}
\paragraph{Is rotation or translation the primary difficulty driver?}
Decomposing unified view distance into rotation and position axes (Figure~\ref{fig:dual_axis}) reveals contrasting difficulty drivers across task types.
P2V/V2P degrade primarily with rotation distance (e.g., GPT-5.4 Pro loses $\sim\!25$ points across rotation bins on P2V), since cumulative rotations are hard to mentally simulate.
IVP reverses this: success collapses with position distance ($\sim\!7\times$ drop for GPT-5.4 Pro), as 3D translation requires spatial layout understanding and path planning beyond simple orientation control.
This contrast aligns with our tracking/planning decomposition: reading a path is largely a matter of \emph{tracking}, which degrades with rotation distance, whereas composing a plan additionally requires reasoning about where an unseen target view lies in the scene layout, a demand that grows with translation rather than orientation.

Sample-level factor analysis (Spearman $\rho$ across $12$ geometric, visual overlap, and directional factors defined in Appendix~\ref{app:factor_definitions}; full results in Appendix~\ref{app:factor_analysis}) further confirms the position bottleneck for IVP and the rotation bottleneck for P2V/V2P.

\vspace{-7pt}
\paragraph{Do successes reflect mental localization or view matching?}

\begin{table}[t]
\centering
\caption{\textbf{Successes reflect view matching, not mental localization.} Each model's successful interactive view planning (IVP)
rollouts, split by whether the agent ever observed a view within the success
threshold ($0.5$\,m / $30^\circ$) of the target view before answering: $\geq\!90\%$ of successes follow such a visual encounter.}
\label{tab:mental_localization}
\small
\setlength{\tabcolsep}{6pt}
\begin{tabular}{@{}lccc@{}}
\toprule
& & \multicolumn{2}{c}{Successful rollouts} \\
\cmidrule(lr){3-4}
Model & \#Success & Visited target view & Inferred (no visit) \\
\midrule
GPT-5.4 Pro & 101 & 99 (98.0\%) & 2 (2.0\%) \\
Gemini 3.1 Pro & 113 & 112 (99.1\%) & 1 (0.9\%) \\
GPT-5.4 & 88 & 83 (94.3\%) & 5 (5.7\%) \\
Grok 4.20 Beta & 42 & 38 (90.5\%) & 4 (9.5\%) \\
Claude Opus 4.6 & 57 & 54 (94.7\%) & 3 (5.3\%) \\
\bottomrule
\end{tabular}
\end{table}

Theoretically, a competent spatial-reasoning model need not see the target view to localize it: after a few informative moves, it could establish an internal spatial mental model, infer how the target view relates to the views it has observed, and submit the target view's camera pose without visiting it.
We test whether current successes of frontier VLMs work this way.
For every successful IVP rollout, we check whether any \emph{observed} camera pose (the initial view plus every view the agent moves to) falls within the success threshold ($0.5$\,m / $30^\circ$) of the target view, \ie{} whether the agent visually encountered the target view before answering (Table~\ref{tab:mental_localization}).
Across all five models, $\geq\!90\%$ of successes (up to $99.1\%$ for Gemini 3.1 Pro) are coupled to such a visual encounter; genuine inference of a correct camera pose without ever visiting a threshold-close view accounts for at most $\sim\!10\%$ of successes.
This suggests that current VLMs mostly succeed by \emph{view matching}, rather than by planning ahead through view space. The gap exposed by IVP is therefore not only a performance gap, but a cognitive gap: today’s models can move through views, but rarely use observed views to forecast and localize a target view that has not yet been fully seen. In short, what they lack is not movement, but planning.

\section{Self-Exploration with View Graph Distillation}
\label{sec:ivet}

Frontier VLMs track local view transitions but cannot compose them to localize an unseen target view. Tracking how a single action changes the view is fundamentally different from planning: composing a sequence of actions whose accumulated observations pin down where an unseen target view lies. We ask whether an agent can bridge this gap through \emph{self-exploration} alone: interacting with the environment, learning from its own experience, and improving without any external demonstration.

Although the action set superficially resembles embodied-navigation primitives, an IVP rollout is at heart a localization problem: actions move only the viewpoint and form a planned trajectory of view manipulations, while reward is granted for an accurate 6-DoF estimate of the target view, not for physically arriving at it. Success thus turns entirely on reasoning: actions serve only to gather evidence for a localization decision made within the turn budget.
This raises a key question: with no external demonstrations and a naive policy succeeding only ${\sim}2.5\%$ of the time, can the agent extract valid supervision from experience that is overwhelmingly unsuccessful?

\subsection{Interactive View Planning as a Localization Problem}
\label{sec:mdp}

We model IVP as a finite-horizon decision process. At each turn $t$, the agent observes the rendered view $o_t$ and its current 6-DoF camera pose $p_t \in \mathrm{SE}(3)$, and selects an action $a_t \in \mathcal{A}$ from the $12$-element action set. The environment updates the pose deterministically, $p_{t+1} = T(p_t, a_t)$, and renders the next view. After at most $T$ turns the agent submits a target estimate $\hat{p}^{*} \in \mathrm{SE}(3)$, scored by
\begin{equation}
  r(\hat{p}^{*}, p^{*}) =
  \mathbf{1}\!\left[\, d_{\mathrm{pos}}(\hat{p}^{*}, p^{*}) \le \beta_t s_t
  \;\wedge\;
  d_{\mathrm{rot}}(\hat{p}^{*}, p^{*}) \le \beta_r s_r \,\right]
  + 0.1\,\mathbf{1}_{\mathrm{format}},
  \label{eq:reward}
\end{equation}
where $p^{*}$ is the ground-truth camera pose of the target view and $\beta_t = \beta_r = 1$ are the human-calibrated thresholds of Section~\ref{sec:data}. A learned policy $\pi_\theta$ maps the rollout history to the next action and, upon termination, to the target estimate.

\subsection{Bootstrapping View Planning from Self-Exploration}
\label{sec:framework}

The key observation is that every trajectory, whether or not it reaches its goal, traces \emph{valid} view transitions through the scene: moving from one viewpoint to another is meaningful supervision regardless of the original target view. Aggregated across exploration, these transitions form a \emph{view graph}, a compact map of how viewpoints connect across a scene, in which connectivity discovered by one episode becomes reusable for any other. This is exactly the any-view-to-any-view structure that composing a plan requires, and it is assembled entirely from the agent's own moves, much as a person builds a spatial map in which even a wrong turn teaches which rooms connect to which hallways. Crucially, there is no stronger teacher here: the agent uncovers the environment's structure by exploring it. In effect, the view graph is an empirical world model of view transitions, holding exactly what the model has experienced; distillation turns it into the model's own.

\begin{figure*}[t]
\centering
\includegraphics[width=\textwidth]{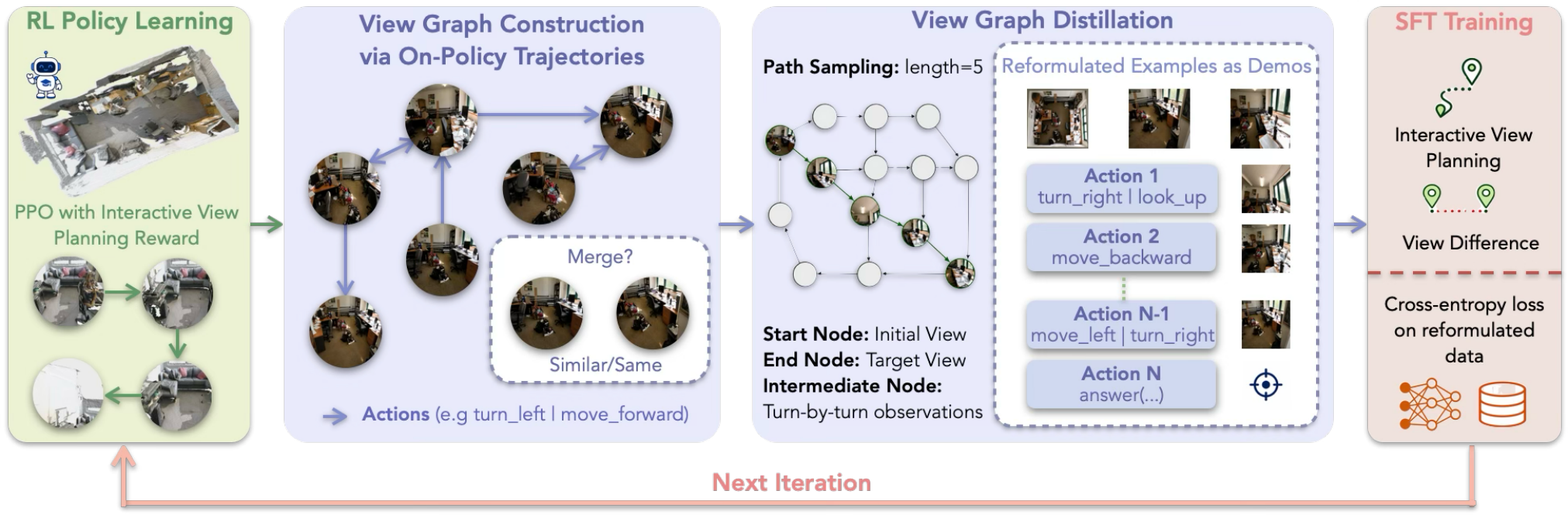}
\caption{Iterative training pipeline. \textbf{Left}: in the \emph{self-exploration} stage, the agent actively explores \viewsuite{} environments under sparse outcome rewards; completed trajectories are continuously compressed into a view graph whose nodes are viewpoints and whose edges are actions. \textbf{Right}: in the \emph{view graph distillation} stage, paths sampled from the graph are reformulated into multi-turn view-planning demonstrations and auxiliary supervision (view-difference estimation and its multiple-choice variant). The distilled model initializes the next self-exploration iteration, progressively bootstrapping the policy.}
\label{fig:ivet_pipeline}
\end{figure*}

We turn this into iterative training that alternates two stages (Figure~\ref{fig:ivet_pipeline}). In each iteration, the agent explores the environment, incrementally compressing its trajectories into the view graph, and a distillation stage samples paths from the graph and reformulates them into supervised view-planning demonstrations that fine-tune the policy. The fine-tuned model initializes the next iteration, so exploration starts from familiar views and pushes outward round by round, growing the graph on-policy toward the views the model will later plan through.

\paragraph{Self-exploration stage.}
The agent interacts with the \viewsuite{} environment using PPO~\citep{ppo} under a sparse reward: an outcome reward of $+1$ when the predicted camera pose of the target view falls within the IVP success threshold ($d_{\mathrm{pos}} \le 0.5\,\mathrm{m}$, $d_{\mathrm{rot}} \le 30^{\circ}$; Section~\ref{sec:data}) and $0$ otherwise, plus a format reward of $+0.1$ for a correctly structured response. A background process continuously converts completed trajectories into the view graph: each node stores a viewpoint with its rendered view, and each directed edge stores the actions taken between two viewpoints. Nodes and edges are deduplicated by viewpoint similarity (Appendix~\ref{app:graph_construction}), so the graph does not grow unboundedly as regions are revisited.

\paragraph{View graph distillation via task reformulation.}
In the distillation stage, we sample paths from the accumulated graph and convert them into supervised data. The mechanism is \emph{task reformulation}. For any path $P = (v_0, a_1, v_1, \ldots, a_K, v_K)$ in the graph, define the operator
\begin{equation}
  \mathcal{R}(P) = \big(\, o_{\mathrm{init}} = v_0,\;\;
  o_{\mathrm{target}} = v_K,\;\;
  (a_1, \ldots, a_K),\;\;
  \hat{p}^{*} = p_{v_K} \,\big),
  \label{eq:reformulation}
\end{equation}
which yields a valid IVP demonstration regardless of whether the original episode succeeded: the end node becomes the target, the start node becomes the initial view, and the action chain becomes the target action sequence. Because $P$ may begin and end at arbitrary nodes, every path is an any-view-to-any-view demonstration, the lever that lets us draw dense supervision from raw, mostly-failed exploration (Algorithm~\ref{alg:ivet} in Appendix~\ref{app:algorithm}). From the same graph we generate three supervision types: (1)~multi-turn view planning via reformulation (the primary task), (2)~view-difference estimation (predicting the unified view distance between two views), and (3)~a multiple-choice variant of view-difference estimation (Appendix~\ref{app:task_reformulation}). The policy is trained with a standard cross-entropy loss using LLaMA-Factory~\citep{llamafactory}; the self-exploration stage is built on VAGEN~\citep{vagen} and veRL~\citep{verl}.

\section{Self-Exploration Closes the Gap}
\label{sec:results_analysis}

\subsection{Experimental Setup}
\label{sec:exp_setup}

We instantiate our framework on two base models: Qwen2.5-VL-7B-Instruct~\citep{qwen25vl} as the primary base, and Qwen3-VL-8B-Instruct as a robustness check. In both cases, iterative training runs for four iterations on $3{,}419$ \viewsuite{} training environments with up to $10$ turns per episode (training/validation split in Appendix~\ref{app:train_val_envs}). The first three iterations alternate a $60$-step self-exploration stage with $3$ epochs of view graph distillation for rapid bootstrapping, and the final iteration runs self-exploration to convergence without further distillation.
Full hyperparameters are provided in Appendix~\ref{app:rl_hyperparams} (RL) and Appendix~\ref{app:sft_hyperparams} (SFT).

\vspace{-7pt}
\paragraph{Prompting Baselines.}
We include the untrained Qwen2.5-VL-7B-Instruct, GPT-5.4 Pro~\citep{gpt54}, and Gemini 3.1 Pro~\citep{gemini31} as zero-shot reference points.

\vspace{-7pt}
\paragraph{Training Baselines.}
We compare against three RL methods, all trained from Qwen2.5-VL-7B-Instruct on the same environments with identical reward and action space:
\begin{itemize}[leftmargin=*,nosep]
\item \textbf{Direct PPO.} PPO~\citep{ppo} training from the base model without any distillation stage. This tests whether self-exploration alone can succeed given sufficient training steps.
\item \textbf{Direct GRPO (filter).} GRPO~\citep{grpo} with $n{=}4$ rollouts per prompt and reward-variance-based filtering~\citep{ragen}. This tests whether an alternative RL algorithm with implicit best-of-$n$ selection can bootstrap learning.
\item \textbf{Success-Only Bootstrapping.} Iterates between PPO and SFT like our framework, but constructs SFT data by filtering successful RL trajectories (reward $> 0.5$) rather than sampling from a view graph with task reformulation. This isolates the contribution of our framework's graph-based data generation from any trajectory, including failures.
\end{itemize}

\vspace{-7pt}
\paragraph{Training Ablations.}
We evaluate three ablations of our framework on Qwen2.5-VL-7B-Instruct:
\begin{itemize}[leftmargin=*,nosep]
\item \textbf{1 iter + RL} and \textbf{2 iter + RL.} Stop after fewer iterations to measure the contribution of the view graph distillation stage.
\item \textbf{Random-graph.} Builds the view graph from a random action generator instead of model-collected trajectories, isolating the contribution of on-policy graph construction.
\end{itemize}

\subsection{Closing the Gap on Interactive View Planning}
\label{sec:main_results_training}

\begin{table}[t]
\centering
\caption{\textbf{Self-exploration with view graph distillation mitigates the planning gap.} Interactive view planning (IVP) success rates (\%) on the \viewsuite{} test set (Short: view distance $d{<}3$; Long: $d{\ge}3$): our framework lifts Qwen2.5-VL-7B from $2.5\%$ to $47.8\%$, above the best frontier model ($21.3\%$), while all training baselines stay below $7\%$ and the Random-graph ablation reaches only $13.0\%$.}
\label{tab:training_results}
\small
\setlength{\tabcolsep}{4pt}
\begin{tabular}{@{}lccc@{}}
\toprule
Method & Short & Long & All \\
\midrule
\multicolumn{4}{@{}l}{\textit{Prompting Baselines}} \\
\midrule
Qwen2.5-VL-7B-Instruct & 7.0 & 0.0 & 2.5 \\
GPT-5.4 Pro & 34.8 & 11.7 & 19.9 \\
Gemini 3.1 Pro & 28.6 & 17.4 & 21.3 \\
\midrule
\multicolumn{4}{@{}l}{\textit{Training Baselines}} \\
\midrule
Direct PPO & 7.0 & 1.2 & 3.2 \\
Direct GRPO (filter) & 10.8 & 2.2 & 5.2 \\
Success-Only Bootstrapping & 14.0 & 2.0 & 6.2 \\
\midrule
\multicolumn{4}{@{}l}{\textit{Training Ablations}} \\
\midrule
Random-graph & 25.4 & 6.4 & 13.0 \\
1 iter + RL & 24.3 & 5.4 & 12.0 \\
2 iter + RL & 49.7 & 16.2 & 27.9 \\
\midrule
\multicolumn{4}{@{}l}{\textbf{\textit{Our Methods}}} \\
\midrule
Qwen2.5-VL-7B-Instruct & 67.2 & 36.9 & 47.8 \\
Qwen3-VL-8B-Instruct & 56.8 & 19.4 & 32.5 \\
\bottomrule
\end{tabular}
\end{table}

As shown in Table~\ref{tab:training_results}, our framework improves Qwen2.5-VL-7B-Instruct from $2.5\%$ to $47.8\%$ on IVP, surpassing all frontier models. Applied to Qwen3-VL-8B-Instruct, the same framework reaches $32.5\%$, above every prompting and training baseline and above the strongest frontier model (Gemini 3.1 Pro, $21.3\%$). The gains thus hold across pre-trained backbones, though their absolute magnitude is backbone-dependent.
All three training baselines remain below $7\%$: Direct PPO ($3.2\%$) confirms that self-exploration alone cannot succeed when the base success rate is near zero; Direct GRPO with filtering ($5.2\%$) shows that best-of-$n$ selection helps only marginally; and Success-Only Bootstrapping ($6.2\%$) underperforms our framework, highlighting the importance of view-graph construction and task reformulation that generate useful supervision from \emph{any} trajectory rather than only successful ones.
Iteration ablations show progressive improvement ($12.0\% \to 27.9\% \to 47.8\%$) across $1$, $2$, and $3$ iterations, while the Random-graph variant achieves only $13.0\%$, confirming that on-policy graph construction is critical, echoing the on-policy advantage of DAgger~\citep{dagger} and on-policy distillation~\citep{onpolicydistill}. Graphs built from random-action trajectories cover regions of view space the model rarely visits during evaluation, so the resulting reformulated supervision transfers poorly. As the policy improves each round, it explores farther and the graph grows outward, covering more of the space the model must plan over.
The ranking between methods is preserved under two evaluation-protocol relaxations, No-Snap (raw rotations executed as-is) and No-Submit (success the moment the pose enters the threshold), so the gain is not an artifact of rotation snapping or of the explicit submit step (Table~\ref{tab:protocol_ablations}, Appendix~\ref{app:protocol_ablations}).

\subsection{What Has the Model Learned?}
\label{sec:analysis}

\paragraph{What exploration strategy does the trained model learn?}
\label{sec:behavior}
\begin{figure*}[t]
\centering
\includegraphics[width=\textwidth]{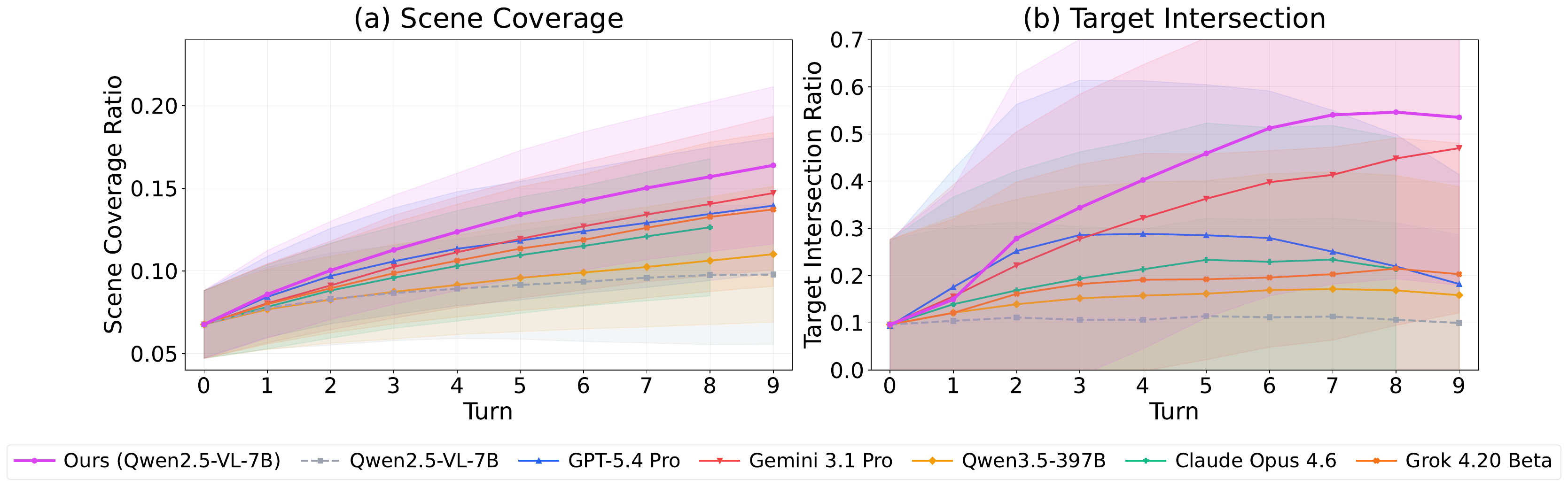}
\caption{\textbf{The trained model learns a two-phase explore-then-approach strategy.} Point-cloud coverage across interactive view planning (IVP) turns, averaged over the test set. \textbf{(a)}~\emph{Scene coverage ratio} (fraction of all scene vertices observed) rises steadily as the agent surveys the scene broadly. \textbf{(b)}~\emph{Target intersection ratio} (fraction of target view vertices covered) grows slowly at first, then accelerates through the middle turns as the agent approaches the target view, an explore-then-approach pattern unique to our trained model; the base and frontier models keep target coverage flat or erratic throughout (full comparison in Appendix~\ref{app:coverage_all}).}
\label{fig:coverage_analysis}
\end{figure*}

We measure 3D point cloud coverage across turns: \textbf{target intersection ratio} (fraction of target view vertices covered) and \textbf{scene coverage ratio} (fraction of all scene vertices observed); see Appendix~\ref{app:coverage_all} for details. Our trained model learns an effective exploration policy (Figure~\ref{fig:coverage_analysis}): scene coverage grows rapidly in early turns as the agent explores broadly, then plateaus; target intersection ratio accelerates in the middle turns as the agent moves toward the target view, reaching $\sim\!55\%$.
This two-phase pattern (explore then approach) is absent in the base model and frontier models, which show flat or erratic target coverage throughout (full model comparison in Appendix~\ref{app:coverage_all}). However, this behavior also suggests a limitation: our trained model still localizes largely by approaching until the target view becomes observable, rather than through prospective spatial reasoning that would localize it beforehand.
\paragraph{How does training reshape the model's attention?}
We also analyze how training changes the model's attention mechanism (Figure~\ref{fig:attention_all_layers}; methodology in Appendix~\ref{app:attention_method}). The trained model allocates more attention to image tokens than the base model across most layers and turns. This suggests that view planning training changes how the model uses observations: instead of treating each view as a weak contextual cue, the model increasingly grounds its decisions in visual evidence accumulated over interaction. The rise of image attention in early and middle turns is consistent with active evidence gathering, while the drop near the final turn suggests a transition from visual grounding to decision formation.

\begin{figure*}[t]
\centering
\includegraphics[width=\textwidth]{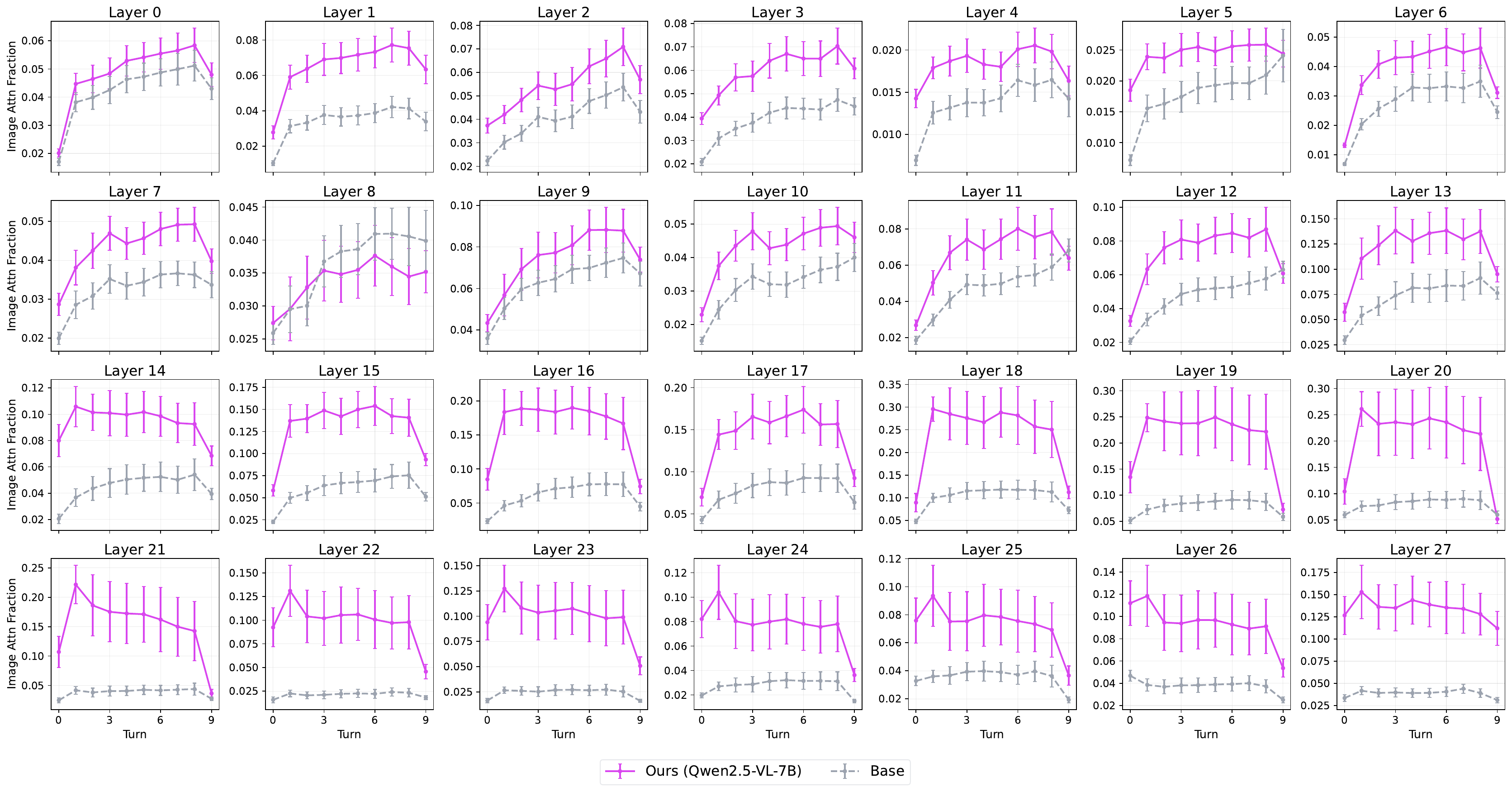}
\caption{\textbf{Training shifts attention toward visual evidence.} Image attention fraction (share of response-token attention on image tokens) across all $28$ layers of trained and base Qwen2.5-VL-7B: the trained model attends to image tokens more across most layers and turns, with the largest gap in the mid-to-late layers; image attention rises in the early and middle turns and falls at the final turn.}
\label{fig:attention_all_layers}
\end{figure*}

\paragraph{Do the learned priors transfer to other view-related tasks?}
\label{sec:transfer}
\begin{table}[t]
\centering
\caption{\textbf{View-planning priors transfer.} Under identical GRPO post-training, our model outperforms the base by $8$--$12$ points on P2V/V2P (transfer within view understanding) and $\sim\!10$ points on MindCube (transfer to an external benchmark), despite similar or lower starting points. Init/Post: accuracy before/after post-training; Base: Qwen2.5-VL-7B-Instruct; Ours: our IVP-trained model.}
\label{tab:downstream_transfer}
\small
\setlength{\tabcolsep}{5pt}
\begin{tabular}{@{}l cc cc cc@{}}
\toprule
& \multicolumn{2}{c}{P2V} & \multicolumn{2}{c}{V2P} & \multicolumn{2}{c}{MindCube} \\
\cmidrule(lr){2-3} \cmidrule(lr){4-5} \cmidrule(lr){6-7}
Model & Init & Post & Init & Post & Init & Post \\
\midrule
Base & 32.1 & 45.1 & 29.2 & 44.8 & 33.0 & 56.3 \\
\rowcolor{gray!10}
Ours & 25.7 & \textbf{57.3} & 31.6 & \textbf{52.8} & 33.1 & \textbf{66.2} \\
\bottomrule
\end{tabular}
\end{table}

We ask whether the spatial priors acquired through interactive view planning transfer to other view-related tasks under further fine-tuning. We test this under identical GRPO post-training on (i)~P2V and V2P from \viewsuite{}, which share scenes and action space with IVP but require different reasoning, and (ii)~MindCube~\citep{mindcube}, an external benchmark with no shared scenes, actions, or rendering pipeline. Details are provided in Appendix~\ref{app:downstream_transfer}.
Despite a slightly lower starting point, our trained model outperforms the base by $8\text{--}12$ points on P2V and V2P after post-training (Table~\ref{tab:downstream_transfer}), indicating that view-planning experience yields priors that go beyond the IVP task itself. On the external MindCube benchmark, our model gains $\sim\!10$ points over the base, showing that the priors transfer to view-dependent spatial reasoning even outside our environment. Interactive view planning is therefore not a narrow skill: it yields spatial priors that strengthen view understanding both within and beyond \viewsuite{}.

\section{Related Work}
\label{sec:related_work}

\vspace{-7pt}
\paragraph{View reasoning and 3D scene benchmarks.}
View-centric QA benchmarks such as MindCube~\citep{mindcube} and ViewSpatial-Bench~\citep{viewspatialbench} probe view-dependent reasoning from images but are non-interactive, as are broader static spatial-QA benchmarks~\citep{spatialrgpt,spatialvlm,3dsrbench} and 3D scene QA built on real scans~\citep{scanqa,sqa3d,threedllm}. Multi-image and video benchmarks~\citep{vsibench,camerabench,dsibench,mmsibench,allangles} add cross-frame and temporal reasoning, and recent work pushes spatial supersensing~\citep{cambrians} and pose-grounded video understanding~\citep{cambrianp} further, yet all treat the model as a passive observer of given frames. Embodied QA and embodied-agent benchmarks~\citep{embodiedbench,hm3dovon,openeqa} introduce active interaction, but optimize for physical arrival at a semantic goal such as an object or a room, where success turns on affordance and traversability. The closest benchmark to ours is E3VS-Bench~\citep{e3vsbench}, which studies viewpoint-dependent active perception in Gaussian-splatting scenes; \viewsuite{} differs in providing full $6$-DoF control, a multi-turn planning task, and a training framework. More broadly, \viewsuite{} targets spatial \emph{localization} through active view planning: the agent plans view manipulations to gather visual evidence, then submits a $6$-DoF estimate of where the target view was taken. Success requires localization accuracy, not physical arrival, isolating viewpoint reasoning from embodied navigation. This framing echoes spatial cognition: humans build cognitive maps of their environment~\citep{tolman1948,okeefe1978}, combine egocentric views with allocentric representations~\citep{burgess2006}, and mentally rotate what they see to anticipate unseen appearances~\citep{shepard1971}, reasoning ahead in the manner of prospective cognition~\citep{schacter2007}. Recent benchmarks begin to ask the same questions of models, probing whether spatial cognition emerges in frontier models~\citep{space}, whether LLMs build cognitive maps that support planning~\citep{cogeval}, and whether VLMs hold a coherent theory of space~\citep{theoryofspace}, form spatial mental models from limited views~\citep{mindcube}, or have internal world models of embodied interaction~\citep{wmabench,enact,embodiedagentinterface}; view planning operationalizes this prospective spatial cognition as a measurable task.

\vspace{-7pt}
\paragraph{Visual search and active perception.}
The closest prior work probes view planning through visual search, an instance of active perception~\citep{bajcsy1988active,aloimonos1988active,bajcsy2018revisiting}. ActiView~\citep{actiview} restricts the action space to zoom and shift within a $2$D image; V$^\star$~\citep{vstar} performs LLM-guided search inside a single high-resolution image; and H$^\star$Bench~\citep{thinking360} studies head rotation over a $360^{\circ}$ panorama. A recent line of agentic visual search trains VLMs to crop, zoom, and reason in pixel space~\citep{minio3,deepeyes,pixelreasoner,chainoffocus,argus}, while embodied agents learn to explore $3$D scenes and decide when to stop~\citep{exploreuntilconfident,threedmem,activeneuralslam}. Because our success metric is a $6$-DoF pose estimate, our task also connects to learning-based camera-pose and geometry estimation~\citep{dust3r,vggt,colmap}, where VLMs are known to struggle~\citep{lostinspace}. \viewsuite{} extends the visual-search thread to real $3$D scenes with full $6$-DoF camera pose control, where the agent must compose a multi-turn plan to localize a target view rather than crop or rotate within a fixed vantage (Table~\ref{tab:benchmark_comparison}).

\vspace{-7pt}
\paragraph{Agentic RL and learning from failure.}
Outcome-supervised RL substantially improves LLM reasoning~\citep{grpo,deepseekr1}, and follow-up work extends it to agentic and multi-modal settings~\citep{zheng2025easyr1,verl,ragen,vagen}. A parallel line of self-improvement bootstraps a policy from its own generations by fine-tuning on successful trajectories~\citep{star,restem,rest,vstar_verifier,rise,sil,spear,relift}. Distilling supervision from a model's own on-policy rollouts, rather than from a fixed off-policy dataset, is also central to on-policy distillation and online imitation learning~\citep{dagger,policydistill,onpolicydistill,lu2025onpolicydistillation}; our view graph is assembled entirely from such on-policy exploration, and we confirm that on-policy construction is critical (Section~\ref{sec:results_analysis}). In classical RL, Hindsight Experience Replay~\citep{her,zhang2023wisdomhindsightmakeslanguage} relabels failed trajectories with the goals they happened to achieve, and recent work adapts this to LM agents by rewriting failed rollouts into supervised targets~\citep{echo}. Our framework builds on this idea, since relabeling a path's endpoint as its target is one instance of our reformulation operator $\mathcal{R}$ (Eq.~\ref{eq:reformulation}), but its aim is broader. Rather than densifying reward episode by episode, we aggregate \emph{all} rollouts into a view graph that captures how viewpoints connect across a scene, and distill this any-view-to-any-view structure into diverse supervised reformulations: multi-turn view planning, view-difference estimation, and its multiple-choice variant. Distilling these reformulations reshapes the policy distribution so that subsequent RL rollouts reach high-reward trajectories more often, combining the distribution sharpening of RL with the distribution reshaping of SFT to overcome the sparse reward under which pure RL fails.

\section{Conclusion and Limitations}
\label{sec:conclusion}

We study \emph{view planning}: composing camera actions into multi-turn plans that localize an unseen target view. Across frontier VLMs, we reveal a clear planning gap: models can track local view transitions, but struggle to compose them into plans toward a target view before it becomes observable. We mitigate this gap with an iterative framework alternating \emph{self-exploration} and \emph{view graph distillation}, improving Qwen2.5-VL-7B from $2.5\%$ to $47.8\%$ on interactive view planning, surpassing all evaluated frontier models, and inducing spatial priors that transfer to related view-understanding tasks. 

\paragraph{Limitations.}
We study static indoor scenes through a discrete $12$-action interface, and validate the framework on two $7$--$8$B backbones; outdoor or dynamic environments, continuous control, and larger model scales are natural next steps. More fundamentally, our trained model still localizes largely by \emph{approaching until the target view becomes observable}, but localizing a target view without ever seeing it, mentally inferring where it lies, remains open. Finally, our framework places little weight on explicit reasoning traces; at our model scale we found it hard to acquire effective explicit reasoning without external expert demonstrations, leaving reasoning-aware training as a further direction.


\subsubsection*{Acknowledgements}
We acknowledge and disclose the use of AI tools in coding and paper writing. We also thank Jiajun Liu, Baiqiao Yin, Jiawei Gu, and Jihan Yang for insightful discussions.

{
\small
\bibliographystyle{plainnat}
\bibliography{references}

@misc{zheng2025easyr1,
  title        = {EasyR1: An Efficient, Scalable, Multi-Modality RL Training Framework},
  author       = {Yaowei Zheng and Junting Lu and Shenzhi Wang and Zhangchi Feng and Dongdong Kuang and Yuwen Xiong and Richong Zhang},
  howpublished = {\url{https://github.com/hiyouga/EasyR1}},
  year         = {2025}
}

@misc{zhang2023wisdomhindsightmakeslanguage,
      title={The Wisdom of Hindsight Makes Language Models Better Instruction Followers}, 
      author={Tianjun Zhang and Fangchen Liu and Justin Wong and Pieter Abbeel and Joseph E. Gonzalez},
      year={2023},
      eprint={2302.05206},
      archivePrefix={arXiv},
      primaryClass={cs.CL},
      url={https://arxiv.org/abs/2302.05206}, 
}

@article{vagen,
  title={VAGEN: Reinforcing World Model Reasoning for Multi-Turn VLM Agents},
  author={Kangrui Wang and Pingyue Zhang and Zihan Wang and Yaning Gao and Linjie Li and Qineng Wang and Hanyang Chen and Chi Wan and Yiping Lu and Zhengyuan Yang and Lijuan Wang and Ranjay Krishna and Jiajun Wu and Li Fei-Fei and Yejin Choi and Manling Li},
  journal={arXiv preprint arXiv:2510.16907},
  year={2025}
}

@article{verl,
  title={HybridFlow: A Flexible and Efficient RLHF Framework},
  author={Guangming Sheng and Chi Zhang and Zilingfeng Ye and Xibin Wu and Wang Zhang and Ru Zhang and Yanghua Peng and Haibin Lin and Chuan Wu},
  journal={arXiv preprint arXiv:2409.19256},
  year={2024}
}

@article{llamafactory,
  title={LlamaFactory: Unified Efficient Fine-Tuning of 100+ Language Models},
  author={Zheng, Yaowei and Zhang, Richong and Zhang, Junhao and Ye, Yanhan and Luo, Zheyan and Ma, Zhangchi and Ma, Yongqiang},
  journal={arXiv preprint arXiv:2403.13372},
  year={2024}
}

@article{ragen,
  title={RAGEN: Understanding Self-Evolution in LLM Agents via Multi-Turn Reinforcement Learning},
  author={Zihan Wang and Kangrui Wang and Qineng Wang and Pingyue Zhang and Linjie Li and Zhengyuan Yang and Xing Jin and Kefan Yu and Minh Nhat Nguyen and Licheng Liu and Eli Gottlieb and Yiping Lu and Kyunghyun Cho and Jiajun Wu and Li Fei-Fei and Lijuan Wang and Yejin Choi and Manling Li},
  journal={arXiv preprint arXiv:2504.20073},
  year={2025}
}

@article{ppo,
  title={Proximal Policy Optimization Algorithms},
  author={Schulman, John and Wolski, Filip and Dhariwal, Prafulla and Radford, Alec and Klimov, Oleg},
  journal={arXiv preprint arXiv:1707.06347},
  year={2017}
}

@inproceedings{scannet,
  author = {Angela Dai and Angel X. Chang and Manolis Savva and Maciej Halber and Thomas A. Funkhouser and Matthias Nie{\ss}ner},
  title = {ScanNet: Richly-Annotated 3D Reconstructions of Indoor Scenes},
  booktitle = {Proceedings of the IEEE/CVF Conference on Computer Vision and Pattern Recognition (CVPR)},
  year = {2017},
  url = {https://doi.org/10.1109/CVPR.2017.261}
}

@inproceedings{scenesplat,
  author = {Li, Yue and Ma, Qi and Yang, Runyi and Li, Huapeng and Ma, Mengjiao and Ren, Bin and Popovic, Nikola and Sebe, Nicu and Konukoglu, Ender and Gevers, Theo and others},
  title = {SceneSplat: Gaussian Splatting-based Scene Understanding with Vision-Language Pretraining},
  booktitle = {Proceedings of the IEEE/CVF International Conference on Computer Vision (ICCV)},
  year = {2025},
  url = {https://arxiv.org/abs/2503.18052}
}

@article{open3d,
  author = {Qian{-}Yi Zhou and Jaesik Park and Vladlen Koltun},
  title = {Open3D: {A} Modern Library for 3D Data Processing},
  journal = {arXiv preprint arXiv:1801.09847},
  year = {2018},
  url = {http://arxiv.org/abs/1801.09847}
}

@article{qwen25vl,
  author = {Shuai Bai and Keqin Chen and Xuejing Liu and Jialin Wang and Wenbin Ge and Sibo Song and Kai Dang and Peng Wang and Shijie Wang and Jun Tang and others},
  title = {Qwen2.5-VL Technical Report},
  journal = {arXiv preprint arXiv:2502.13923},
  year = {2025},
  url = {https://arxiv.org/abs/2502.13923}
}

@misc{grpo,
      title={DeepSeekMath: Pushing the Limits of Mathematical Reasoning in Open Language Models}, 
      author={Zhihong Shao and Peiyi Wang and Qihao Zhu and Runxin Xu and Junxiao Song and Xiao Bi and Haowei Zhang and Mingchuan Zhang and Y. K. Li and Y. Wu and Daya Guo},
      year={2024},
      eprint={2402.03300},
      archivePrefix={arXiv},
      primaryClass={cs.CL},
      url={https://arxiv.org/abs/2402.03300}, 
}

@article{deepseekr1,
  author = {DeepSeek{-}AI},
  title = {DeepSeek-R1: Incentivizing Reasoning Capability in LLMs via Reinforcement Learning},
  journal = {arXiv preprint arXiv:2501.12948},
  year = {2025},
  url = {https://arxiv.org/abs/2501.12948}
}

@inproceedings{her,
  author = {Andrychowicz, Marcin and Wolski, Filip and Ray, Alex and Schneider, Jonas and Fong, Rachel and Welinder, Peter and McGrew, Bob and Tobin, Josh and Abbeel, Pieter and Zaremba, Wojciech},
  title = {Hindsight Experience Replay},
  booktitle = {Advances in Neural Information Processing Systems},
  year = {2017}
}

@article{gpt5,
  author = {{OpenAI}},
  title = {{GPT-5} System Card},
  journal = {arXiv preprint arXiv:2601.03267},
  year = {2026},
  url = {https://arxiv.org/abs/2601.03267}
}

@misc{gpt54,
  author = {{OpenAI}},
  title = {{GPT-5.4} Thinking System Card},
  year = {2026},
  url = {https://deploymentsafety.openai.com/gpt-5-4-thinking}
}

@misc{gemini3,
  author = {{Google DeepMind}},
  title = {Gemini 3 Pro Model Card},
  year = {2025},
  url = {https://deepmind.google/models/model-cards/gemini-3-pro/}
}

@misc{gemini31,
  author = {{Google DeepMind}},
  title = {Gemini 3.1 Pro Model Card},
  year = {2026},
  url = {https://deepmind.google/models/model-cards/gemini-3-1-pro/}
}

@misc{claude46,
  author = {{Anthropic}},
  title = {Claude Opus 4.6 System Card},
  year = {2026},
  url = {https://www.anthropic.com/claude-opus-4-6-system-card}
}

@article{glmv,
  author = {Wenyi Hong and others},
  title = {{GLM-4.5V} and {GLM-4.1V-Thinking}: Towards Versatile Multimodal Reasoning with Scalable Reinforcement Learning},
  journal = {arXiv preprint arXiv:2507.01006},
  year = {2025},
  url = {https://arxiv.org/abs/2507.01006}
}

@inproceedings{spatialrgpt,
  author = {An{-}Chieh Cheng and Hongxu Yin and Yang Fu and Qiushan Guo and Ruihan Yang and Jan Kautz and Xiaolong Wang and Sifei Liu},
  title = {SpatialRGPT: Grounded Spatial Reasoning in Vision-Language Models},
  booktitle = {Advances in Neural Information Processing Systems (NeurIPS)},
  year = {2024},
  url = {http://papers.nips.cc/paper_files/paper/2024/hash/f38cb4cf9a5eaa92b3cfa481832719c6-Abstract-Conference.html}
}

@misc{spatialvlm,
      title={SpatialVLM: Endowing Vision-Language Models with Spatial Reasoning Capabilities}, 
      author={Boyuan Chen and Zhuo Xu and Sean Kirmani and Brian Ichter and Danny Driess and Pete Florence and Dorsa Sadigh and Leonidas Guibas and Fei Xia},
      year={2024},
      eprint={2401.12168},
      archivePrefix={arXiv},
      primaryClass={cs.CV},
      url={https://arxiv.org/abs/2401.12168}, 
}

@inproceedings{vsibench,
  author = {Jihan Yang and Shusheng Yang and Anjali W. Gupta and Rilyn Han and Li Fei{-}Fei and Saining Xie},
  title = {Thinking in Space: How Multimodal Large Language Models See, Remember, and Recall Spaces},
  booktitle = {Proceedings of the IEEE/CVF Conference on Computer Vision and Pattern Recognition (CVPR)},
  year = {2025},
  url = {https://doi.org/10.1109/CVPR52734.2025.00994}
}

@article{camerabench,
  author = {Zhiqiu Lin and Siyuan Cen and Daniel Jiang and Jay Karhade and Hewei Wang and Chancharik Mitra and Tiffany Ling and Yuhan Huang and Sifan Liu and Mingyu Chen and Rushikesh Zawar and Xue Bai and Yilun Du and Chuang Gan and Deva Ramanan},
  title = {Towards Understanding Camera Motions in Any Video},
  journal = {arXiv preprint arXiv:2504.15376},
  year = {2025},
  url = {https://arxiv.org/abs/2504.15376}
}

@misc{dsibench,
      title={DSI-Bench: A Benchmark for Dynamic Spatial Intelligence}, 
      author={Ziang Zhang and Zehan Wang and Guanghao Zhang and Weilong Dai and Yan Xia and Ziang Yan and Minjie Hong and Zhou Zhao},
      year={2025},
      eprint={2510.18873},
      archivePrefix={arXiv},
      primaryClass={cs.CV},
      url={https://arxiv.org/abs/2510.18873}, 
}

@misc{3dsrbench,
      title={3DSRBench: A Comprehensive 3D Spatial Reasoning Benchmark}, 
      author={Wufei Ma and Haoyu Chen and Guofeng Zhang and Yu-Cheng Chou and Jieneng Chen and Celso M de Melo and Alan Yuille},
      year={2025},
      eprint={2412.07825},
      archivePrefix={arXiv},
      primaryClass={cs.CV},
      url={https://arxiv.org/abs/2412.07825}, 
}

@article{mindcube,
  author = {Qineng Wang and Baiqiao Yin and Pingyue Zhang and Jianshu Zhang and Kangrui Wang and Zihan Wang and Jieyu Zhang and Keshigeyan Chandrasegaran and Han Liu and Ranjay Krishna and Saining Xie and Manling Li and Jiajun Wu and Li Fei{-}Fei},
  title = {Spatial Mental Modeling from Limited Views},
  journal = {arXiv preprint arXiv:2506.21458},
  year = {2025},
  url = {https://arxiv.org/abs/2506.21458}
}

@inproceedings{embodiedbench,
  author = {Rui Yang and Hanyang Chen and Junyu Zhang and Mark Zhao and Cheng Qian and Kangrui Wang and Qineng Wang and Teja Venkat Koripella and Marziyeh Movahedi and Manling Li and Heng Ji and Huan Zhang and Tong Zhang},
  title = {EmbodiedBench: Comprehensive Benchmarking Multi-modal Large Language Models for Vision-Driven Embodied Agents},
  booktitle = {Proceedings of the International Conference on Machine Learning (ICML)},
  year = {2025},
  url = {https://proceedings.mlr.press/v267/yang25f.html}
}

@inproceedings{theoryofspace,
  author = {Pingyue Zhang and Zihan Huang and Yue Wang and Jieyu Zhang and Letian Xue and Zihan Wang and Qineng Wang and Keshigeyan Chandrasegaran and Ruohan Zhang and Yejin Choi and Ranjay Krishna and Jiajun Wu and Li Fei{-}Fei and Manling Li},
  title = {Theory of Space: Can Foundation Models Construct Spatial Beliefs through Active Exploration?},
  booktitle = {Proceedings of the International Conference on Learning Representations (ICLR)},
  year = {2026},
  url = {https://arxiv.org/abs/2602.07055}
}

@article{viewspatialbench,
  author = {Dingming Li and Hongxing Li and Zixuan Wang and Yuchen Yan and Hang Zhang and Siqi Chen and Guiyang Hou and Shengpei Jiang and Wenqiao Zhang and Yongliang Shen and Weiming Lu and Yueting Zhuang},
  title = {ViewSpatial-Bench: Evaluating Multi-perspective Spatial Localization in Vision-Language Models},
  journal = {arXiv preprint arXiv:2505.21500},
  year = {2025},
  url = {https://arxiv.org/abs/2505.21500}
}

@article{bajcsy1988active,
  author={Bajcsy, R.},
  journal={Proceedings of the IEEE}, 
  title={Active perception}, 
  year={1988},
}

@article{aloimonos1988active,
  author = {Aloimonos, John and Weiss, Isaac and Bandyopadhyay, Amit},
  title = {Active Vision},
  journal = {International Journal of Computer Vision},
  year = {1988}
}

@article{bajcsy2018revisiting,
  author = {Bajcsy, Ruzena and Aloimonos, Yiannis and Tsotsos, John K.},
  title = {Revisiting Active Perception},
  journal = {Autonomous Robots},
  volume = {42},
  number = {2},
  pages = {177--196},
  year = {2018},
  url = {https://arxiv.org/abs/1603.02729}
}

@misc{actiview,
      title={ActiView: Evaluating Active Perception Ability for Multimodal Large Language Models}, 
      author={Ziyue Wang and Chi Chen and Fuwen Luo and Yurui Dong and Yuanchi Zhang and Yuzhuang Xu and Xiaolong Wang and Peng Li and Yang Liu},
      year={2025},
      eprint={2410.04659},
      archivePrefix={arXiv},
      primaryClass={cs.CV},
      url={https://arxiv.org/abs/2410.04659}, 
}

@inproceedings{vstar,
  author = {Wu, Penghao and Xie, Saining},
  title = {{V*}: Guided Visual Search as a Core Mechanism in Multimodal {LLMs}},
  booktitle = {Proceedings of the IEEE/CVF Conference on Computer Vision and Pattern Recognition (CVPR)},
  year = {2024},
  url = {https://arxiv.org/abs/2312.14135}
}

@misc{thinking360,
      title={Thinking in 360{\textdegree}: Humanoid Visual Search in the Wild},
      author={Heyang Yu and Yinan Han and Xiangyu Zhang and Baiqiao Yin and Bowen Chang and Xiangyu Han and Xinhao Liu and Jing Zhang and Marco Pavone and Chen Feng and Saining Xie and Yiming Li},
      year={2025},
      eprint={2511.20351},
      archivePrefix={arXiv},
      primaryClass={cs.CV},
      url={https://arxiv.org/abs/2511.20351}, 
}

@article{hm3dovon,
  author = {Yokoyama, Naoki and Ramrakhya, Ram and Das, Abhishek and Batra, Dhruv and Ha, Sehoon},
  title = {{HM3D-OVON}: A Dataset and Benchmark for Open-Vocabulary Object Goal Navigation},
  journal = {arXiv preprint arXiv:2409.14296},
  year = {2024},
  url = {https://arxiv.org/abs/2409.14296}
}

@misc{gsplat,
      title={3D Gaussian Splatting for Real-Time Radiance Field Rendering}, 
      author={Bernhard Kerbl and Georgios Kopanas and Thomas Leimkühler and George Drettakis},
      year={2023},
      eprint={2308.04079},
      archivePrefix={arXiv},
      primaryClass={cs.GR},
      url={https://arxiv.org/abs/2308.04079}, 
}

@article{e3vsbench,
  author = {Koya Sakamoto and Taiki Miyanishi and Daichi Azuma and Shuhei Kurita and Shu Morikuni and Naoya Chiba and Motoaki Kawanabe and Yusuke Iwasawa and Yutaka Matsuo},
  title = {E3VS-Bench: A Benchmark for Viewpoint-Dependent Active Perception in 3D Gaussian Splatting Scenes},
  journal = {arXiv preprint arXiv:2604.17969},
  year = {2026},
  url = {https://arxiv.org/abs/2604.17969}
}

@inproceedings{scanqa,
  author = {Daichi Azuma and Taiki Miyanishi and Shuhei Kurita and Motoaki Kawanabe},
  title = {ScanQA: 3D Question Answering for Spatial Scene Understanding},
  booktitle = {Proceedings of the IEEE/CVF Conference on Computer Vision and Pattern Recognition (CVPR)},
  year = {2022},
  url = {https://arxiv.org/abs/2112.10482}
}

@inproceedings{sqa3d,
  author = {Xiaojian Ma and Silong Yong and Zilong Zheng and Qing Li and Yitao Liang and Song-Chun Zhu and Siyuan Huang},
  title = {SQA3D: Situated Question Answering in 3D Scenes},
  booktitle = {International Conference on Learning Representations (ICLR)},
  year = {2023},
  url = {https://arxiv.org/abs/2210.07474}
}

@inproceedings{threedllm,
  author = {Yining Hong and Haoyu Zhen and Peihao Chen and Shuhong Zheng and Yilun Du and Zhenfang Chen and Chuang Gan},
  title = {3D-LLM: Injecting the 3D World into Large Language Models},
  booktitle = {Advances in Neural Information Processing Systems (NeurIPS)},
  year = {2023},
  url = {https://arxiv.org/abs/2307.12981}
}

@inproceedings{mmsibench,
  author = {Sihan Yang and Runsen Xu and Yiman Xie and Sizhe Yang and Mo Li and Jingli Lin and Chenming Zhu and Xiaochen Chen and Haodong Duan and Xiangyu Yue and Dahua Lin and Tai Wang and Jiangmiao Pang},
  title = {MMSI-Bench: A Benchmark for Multi-Image Spatial Intelligence},
  booktitle = {International Conference on Learning Representations (ICLR)},
  year = {2026},
  url = {https://arxiv.org/abs/2505.23764}
}

@article{allangles,
  author = {Chun-Hsiao Yeh and Chenyu Wang and Shengbang Tong and Ta-Ying Cheng and Ruoyu Wang and Tianzhe Chu and Yuexiang Zhai and Yubei Chen and Shenghua Gao and Yi Ma},
  title = {Seeing from Another Perspective: Evaluating Multi-View Understanding in MLLMs},
  journal = {arXiv preprint arXiv:2504.15280},
  year = {2025},
  url = {https://arxiv.org/abs/2504.15280}
}

@article{cambrians,
  author = {Shusheng Yang and Jihan Yang and Pinzhi Huang and Ellis Brown and Zihao Yang and Yue Yu and Shengbang Tong and Zihan Zheng and Yifan Xu and Muhan Wang and Daohan Lu and Rob Fergus and Yann LeCun and Li Fei-Fei and Saining Xie},
  title = {Cambrian-S: Towards Spatial Supersensing in Video},
  journal = {arXiv preprint arXiv:2511.04670},
  year = {2025},
  url = {https://arxiv.org/abs/2511.04670}
}

@inproceedings{openeqa,
  author = {Arjun Majumdar and Anurag Ajay and Xiaohan Zhang and Pranav Putta and Sriram Yenamandra and Mikael Henaff and Sneha Silwal and Paul McVay and Oleksandr Maksymets and Sergio Arnaud and Karmesh Yadav and Qiyang Li and Ben Newman and Mohit Sharma and Vincent Berges and Shiqi Zhang and Pulkit Agrawal and Yonatan Bisk and Dhruv Batra and Mrinal Kalakrishnan and Franziska Meier and Chris Paxton and Alexander Sax and Aravind Rajeswaran},
  title = {OpenEQA: Embodied Question Answering in the Era of Foundation Models},
  booktitle = {Proceedings of the IEEE/CVF Conference on Computer Vision and Pattern Recognition (CVPR)},
  year = {2024},
  url = {https://openaccess.thecvf.com/content/CVPR2024/html/Majumdar_OpenEQA_Embodied_Question_Answering_in_the_Era_of_Foundation_Models_CVPR_2024_paper.html}
}

@inproceedings{dust3r,
  author = {Shuzhe Wang and Vincent Leroy and Yohann Cabon and Boris Chidlovskii and Jerome Revaud},
  title = {DUSt3R: Geometric 3D Vision Made Easy},
  booktitle = {Proceedings of the IEEE/CVF Conference on Computer Vision and Pattern Recognition (CVPR)},
  year = {2024},
  url = {https://arxiv.org/abs/2312.14132}
}

@inproceedings{vggt,
  author = {Jianyuan Wang and Minghao Chen and Nikita Karaev and Andrea Vedaldi and Christian Rupprecht and David Novotny},
  title = {VGGT: Visual Geometry Grounded Transformer},
  booktitle = {Proceedings of the IEEE/CVF Conference on Computer Vision and Pattern Recognition (CVPR)},
  year = {2025},
  url = {https://arxiv.org/abs/2503.11651}
}

@article{lostinspace,
  author = {Ken Deng and Yifu Qiu and Yoni Kasten and Shay B. Cohen and Yftah Ziser},
  title = {Lost in Space? Vision-Language Models Struggle with Relative Camera Pose Estimation},
  journal = {arXiv preprint arXiv:2601.22228},
  year = {2026},
  url = {https://arxiv.org/abs/2601.22228}
}

@inproceedings{colmap,
  author = {Johannes L. Sch\"{o}nberger and Jan-Michael Frahm},
  title = {Structure-from-Motion Revisited},
  booktitle = {Proceedings of the IEEE/CVF Conference on Computer Vision and Pattern Recognition (CVPR)},
  year = {2016}
}

@article{minio3,
  author = {Xin Lai and Junyi Li and Wei Li and Tao Liu and Tianjian Li and Hengshuang Zhao},
  title = {Mini-o3: Scaling Up Reasoning Patterns and Interaction Turns for Visual Search},
  journal = {arXiv preprint arXiv:2509.07969},
  year = {2025},
  url = {https://arxiv.org/abs/2509.07969}
}

@article{deepeyes,
  author = {Ziwei Zheng and Michael Yang and Jack Hong and Chenxiao Zhao and Guohai Xu and Le Yang and Chao Shen and Xing Yu},
  title = {DeepEyes: Incentivizing Thinking with Images via Reinforcement Learning},
  journal = {arXiv preprint arXiv:2505.14362},
  year = {2025},
  url = {https://arxiv.org/abs/2505.14362}
}

@article{pixelreasoner,
  author = {Haozhe Wang and Alex Su and Weiming Ren and Fangzhen Lin and Wenhu Chen},
  title = {Pixel Reasoner: Incentivizing Pixel-Space Reasoning with Curiosity-Driven Reinforcement Learning},
  journal = {arXiv preprint arXiv:2505.15966},
  year = {2025},
  url = {https://arxiv.org/abs/2505.15966}
}

@article{chainoffocus,
  author = {Xintong Zhang and Zhi Gao and Bofei Zhang and Pengxiang Li and Xiaowen Zhang and Yang Liu and Tao Yuan and Yuwei Wu and Yunde Jia and Song-Chun Zhu and Qing Li},
  title = {Adaptive Chain-of-Focus Reasoning via Dynamic Visual Search and Zooming for Efficient VLMs},
  journal = {arXiv preprint arXiv:2505.15436},
  year = {2025},
  url = {https://arxiv.org/abs/2505.15436}
}

@inproceedings{argus,
  author = {Yunze Man and De-An Huang and Guilin Liu and Shiwei Sheng and Shilong Liu and Liang-Yan Gui and Jan Kautz and Yu-Xiong Wang and Zhiding Yu},
  title = {Argus: Vision-Centric Reasoning with Grounded Chain-of-Thought},
  booktitle = {Proceedings of the IEEE/CVF Conference on Computer Vision and Pattern Recognition (CVPR)},
  year = {2025},
  url = {https://arxiv.org/abs/2505.23766}
}

@inproceedings{exploreuntilconfident,
  author = {Allen Z. Ren and Jaden Clark and Anushri Dixit and Masha Itkina and Anirudha Majumdar and Dorsa Sadigh},
  title = {Explore until Confident: Efficient Exploration for Embodied Question Answering},
  booktitle = {Robotics: Science and Systems (RSS)},
  year = {2024},
  url = {https://arxiv.org/abs/2403.15941}
}

@article{threedmem,
  author = {Yuncong Yang and Han Yang and Jiachen Zhou and Peihao Chen and Hongxin Zhang and Yilun Du and Chuang Gan},
  title = {3D-Mem: 3D Scene Memory for Embodied Exploration and Reasoning},
  journal = {arXiv preprint arXiv:2411.17735},
  year = {2024},
  url = {https://arxiv.org/abs/2411.17735}
}

@inproceedings{activeneuralslam,
  author = {Devendra Singh Chaplot and Dhiraj Gandhi and Saurabh Gupta and Abhinav Gupta and Ruslan Salakhutdinov},
  title = {Learning to Explore using Active Neural SLAM},
  booktitle = {International Conference on Learning Representations (ICLR)},
  year = {2020},
  url = {https://arxiv.org/abs/2004.05155}
}

@inproceedings{star,
  author = {Eric Zelikman and Yuhuai Wu and Jesse Mu and Noah D. Goodman},
  title = {STaR: Bootstrapping Reasoning With Reasoning},
  booktitle = {Advances in Neural Information Processing Systems (NeurIPS)},
  year = {2022},
  url = {https://arxiv.org/abs/2203.14465}
}

@article{restem,
  author = {Avi Singh and John D. Co-Reyes and Rishabh Agarwal and others},
  title = {Beyond Human Data: Scaling Self-Training for Problem-Solving with Language Models},
  journal = {Transactions on Machine Learning Research (TMLR)},
  year = {2024},
  url = {https://arxiv.org/abs/2312.06585}
}

@article{rest,
  author = {Caglar Gulcehre and Tom Le Paine and Srivatsan Srinivasan and Ksenia Konyushkova and Lotte Weerts and Abhishek Sharma and Aditya Siddhant and Alex Ahern and Miaosen Wang and Chenjie Gu and Wolfgang Macherey and Arnaud Doucet and Orhan Firat and Nando de Freitas},
  title = {Reinforced Self-Training (ReST) for Language Modeling},
  journal = {arXiv preprint arXiv:2308.08998},
  year = {2023},
  url = {https://arxiv.org/abs/2308.08998}
}

@inproceedings{vstar_verifier,
  author = {Arian Hosseini and Xingdi Yuan and Nikolay Malkin and Aaron Courville and Alessandro Sordoni and Rishabh Agarwal},
  title = {V-STaR: Training Verifiers for Self-Taught Reasoners},
  booktitle = {Conference on Language Modeling (COLM)},
  year = {2024},
  url = {https://arxiv.org/abs/2402.06457}
}

@inproceedings{rise,
  author = {Yuxiao Qu and Tianjun Zhang and Naman Garg and Aviral Kumar},
  title = {Recursive Introspection: Teaching Language Model Agents How to Self-Improve},
  booktitle = {Advances in Neural Information Processing Systems (NeurIPS)},
  year = {2024},
  url = {https://arxiv.org/abs/2407.18219}
}

@inproceedings{sil,
  author = {Junhyuk Oh and Yijie Guo and Satinder Singh and Honglak Lee},
  title = {Self-Imitation Learning},
  booktitle = {Proceedings of the International Conference on Machine Learning (ICML)},
  year = {2018},
  url = {https://arxiv.org/abs/1806.05635}
}

@article{spear,
  author = {Yulei Qin and Xiaoyu Tan and Zhengbao He and Gang Li and Haojia Lin and Zongyi Li and Zihan Xu and Yuchen Shi and Siqi Cai and Renting Rui and Shaofei Cai and Yuzheng Cai and Xuan Zhang and Sheng Ye and Ke Li and Xing Sun},
  title = {Learn the Ropes, Then Trust the Wins: Self-imitation with Progressive Exploration for Agentic Reinforcement Learning},
  journal = {arXiv preprint arXiv:2509.22601},
  year = {2025},
  url = {https://arxiv.org/abs/2509.22601}
}

@inproceedings{relift,
  author = {Lu Ma and Hao Liang and Meiyi Qiang and Lexiang Tang and Xiaochen Ma and Zhen Hao Wong and Junbo Niu and Chengyu Shen and Runming He and Yanhao Li and Bin Cui and Wentao Zhang},
  title = {Learning What Reinforcement Learning Can't: Interleaved Online Fine-Tuning for Hardest Questions},
  booktitle = {International Conference on Learning Representations (ICLR)},
  year = {2026},
  url = {https://arxiv.org/abs/2506.07527}
}

@article{echo,
  author = {Michael Y. Hu and Benjamin Van Durme and Jacob Andreas and Harsh Jhamtani},
  title = {Sample-Efficient Online Learning in LM Agents via Hindsight Trajectory Rewriting},
  journal = {arXiv preprint arXiv:2510.10304},
  year = {2025},
  url = {https://arxiv.org/abs/2510.10304}
}

@inproceedings{embodiedagentinterface,
  author = {Manling Li and Shiyu Zhao and Qineng Wang and Kangrui Wang and Yu Zhou and Sanjana Srivastava and Cem Gokmen and Tony Lee and Li Erran Li and Ruohan Zhang and Weiyu Liu and Percy Liang and Li Fei-Fei and Jiayuan Mao and Jiajun Wu},
  title = {Embodied Agent Interface: Benchmarking LLMs for Embodied Decision Making},
  booktitle = {Advances in Neural Information Processing Systems (NeurIPS) Datasets and Benchmarks Track},
  year = {2024},
  url = {https://arxiv.org/abs/2410.07166}
}

@misc{cambrianp,
      title={{Cambrian-P}: Pose-Grounded Video Understanding},
      author={Jihan Yang and Zifan Zhao and Xichen Pan and Shusheng Yang and Junyi Zhang and Bingyi Kang and Hu Xu and Saining Xie},
      year={2026},
      eprint={2605.22819},
      archivePrefix={arXiv},
      primaryClass={cs.CV},
      url={https://arxiv.org/abs/2605.22819},
}

@inproceedings{onpolicydistill,
  author = {Rishabh Agarwal and Nino Vieillard and Yongchao Zhou and Piotr Stanczyk and Sabela Ramos and Matthieu Geist and Olivier Bachem},
  title = {On-Policy Distillation of Language Models: Learning from Self-Generated Mistakes},
  booktitle = {International Conference on Learning Representations (ICLR)},
  year = {2024},
  url = {https://arxiv.org/abs/2306.13649}
}

@inproceedings{dagger,
  author = {Stephane Ross and Geoffrey J. Gordon and J. Andrew Bagnell},
  title = {A Reduction of Imitation Learning and Structured Prediction to No-Regret Online Learning},
  booktitle = {Proceedings of the Fourteenth International Conference on Artificial Intelligence and Statistics (AISTATS)},
  year = {2011},
  url = {https://arxiv.org/abs/1011.0686}
}

@article{lu2025onpolicydistillation,
  author = {Kevin Lu and {Thinking Machines Lab}},
  title = {On-Policy Distillation},
  journal = {Thinking Machines Lab: Connectionism},
  year = {2025},
  url = {https://thinkingmachines.ai/blog/on-policy-distillation}
}

@inproceedings{policydistill,
  author = {Andrei A. Rusu and Sergio Gomez Colmenarejo and Caglar Gulcehre and Guillaume Desjardins and James Kirkpatrick and Razvan Pascanu and Volodymyr Mnih and Koray Kavukcuoglu and Raia Hadsell},
  title = {Policy Distillation},
  booktitle = {International Conference on Learning Representations (ICLR)},
  year = {2016},
  url = {https://arxiv.org/abs/1511.06295}
}

@article{enact,
  author = {Qineng Wang and Wenlong Huang and Yu Zhou and Hang Yin and Tianwei Bao and Jianwen Lyu and Weiyu Liu and Ruohan Zhang and Jiajun Wu and Li Fei-Fei and Manling Li},
  title = {{ENACT}: Evaluating Embodied Cognition with World Modeling of Egocentric Interaction},
  journal = {arXiv preprint arXiv:2511.20937},
  year = {2025},
  url = {https://arxiv.org/abs/2511.20937}
}

@article{tolman1948,
  author = {Edward C. Tolman},
  title = {Cognitive Maps in Rats and Men},
  journal = {Psychological Review},
  volume = {55},
  number = {4},
  pages = {189--208},
  year = {1948}
}

@article{shepard1971,
  author = {Roger N. Shepard and Jacqueline Metzler},
  title = {Mental Rotation of Three-Dimensional Objects},
  journal = {Science},
  volume = {171},
  number = {3972},
  pages = {701--703},
  year = {1971}
}

@book{okeefe1978,
  author = {John O'Keefe and Lynn Nadel},
  title = {The Hippocampus as a Cognitive Map},
  publisher = {Oxford University Press},
  year = {1978}
}

@article{schacter2007,
  author = {Daniel L. Schacter and Donna Rose Addis and Randy L. Buckner},
  title = {Remembering the Past to Imagine the Future: The Prospective Brain},
  journal = {Nature Reviews Neuroscience},
  volume = {8},
  pages = {657--661},
  year = {2007}
}

@article{burgess2006,
  author = {Neil Burgess},
  title = {Spatial Memory: How Egocentric and Allocentric Combine},
  journal = {Trends in Cognitive Sciences},
  volume = {10},
  number = {12},
  pages = {551--557},
  year = {2006}
}

@inproceedings{space,
  author = {Santhosh Kumar Ramakrishnan and Erik Wijmans and Philipp Kr{\"a}henb{\"u}hl and Vladlen Koltun},
  title = {Does Spatial Cognition Emerge in Frontier Models?},
  booktitle = {Proceedings of the International Conference on Learning Representations (ICLR)},
  year = {2025},
  url = {https://arxiv.org/abs/2410.06468}
}

@inproceedings{cogeval,
  author = {Ida Momennejad and Hosein Hasanbeig and Felipe Vieira Frujeri and Hiteshi Sharma and Nebojsa Jojic and Hamid Palangi and Robert Ness and Jonathan Larson},
  title = {Evaluating Cognitive Maps and Planning in Large Language Models with {CogEval}},
  booktitle = {Advances in Neural Information Processing Systems (NeurIPS)},
  year = {2023},
  url = {https://arxiv.org/abs/2309.15129}
}

@inproceedings{wmabench,
  author = {Qiyue Gao and Xinyu Pi and Kevin Liu and Junrong Chen and Ruolan Yang and Xinqi Huang and Xinyu Fang and Lu Sun and Gautham Kishore and Bo Ai and Stone Tao and Mengyang Liu and Jiaxi Yang and Chao{-}Jung Lai and Chuanyang Jin and Jiannan Xiang and Benhao Huang and Zeming Chen and David Danks and Hao Su and Tianmin Shu and Ziqiao Ma and Lianhui Qin and Zhiting Hu},
  title = {Do Vision-Language Models Have Internal World Models? Towards an Atomic Evaluation},
  booktitle = {Findings of the Association for Computational Linguistics: ACL 2025},
  year = {2025},
  url = {https://arxiv.org/abs/2506.21876}
}

@misc{grok420,
  author = {{xAI}},
  title = {Grok 4.20 Model Documentation},
  year = {2026},
  url = {https://docs.x.ai/developers/models/grok-4.20}
}

@article{kimik25,
  author = {{Kimi Team}},
  title = {Kimi {K2.5}: Visual Agentic Intelligence},
  journal = {arXiv preprint arXiv:2602.02276},
  year = {2026},
  url = {https://arxiv.org/abs/2602.02276}
}

@article{qwen3vl,
  author = {Shuai Bai and others},
  title = {{Qwen3-VL} Technical Report},
  journal = {arXiv preprint arXiv:2511.21631},
  year = {2025},
  url = {https://arxiv.org/abs/2511.21631}
}

@misc{qwen35,
  author = {{Qwen Team}},
  title = {Qwen3.5: Towards Native Multimodal Agents},
  year = {2026},
  url = {https://qwen.ai/blog?id=qwen3.5}
}
}

\appendix
\newpage
\renewcommand{\partname}{}
\part{Appendix}
\parttoc
\clearpage

\section{\viewsuite{} Details}
\label{app:viewsuite}

\subsection{Action Space Details}
\label{app:action_space}

As described in Section~\ref{sec:environment}, \viewsuite{} provides $12$ camera actions (Figure~\ref{fig:environment_overview}a). Table~\ref{tab:action_space_detail} lists all actions with their geometric definitions.
The camera coordinate frame follows the OpenCV convention: $+X$ is screen-right, $+Y$ is screen-down, and $+Z$ points into the scene (forward).
The world coordinate frame uses the ScanNet convention with $Z$-up.

\paragraph{Translation actions.}
The six translation actions move the camera center along its local axes by $s_t = 0.5$\,m per step.
\texttt{move\_forward} / \texttt{move\_backward} translate along the camera's $+Z$ / $-Z$ axis;
\texttt{move\_left} / \texttt{move\_right} translate along $-X$ / $+X$;
\texttt{move\_up} / \texttt{move\_down} translate along the screen-up / screen-down direction ($-Y$ / $+Y$ under the OpenCV convention).

\paragraph{Rotation actions.}
The six rotation actions rotate the camera about its center by $s_r = 30^\circ$ per step.
\texttt{turn\_left} / \texttt{turn\_right} apply yaw (rotation about the camera's local $Y$ axis);
\texttt{look\_up} / \texttt{look\_down} apply pitch (rotation about the local $X$ axis);
\texttt{rotate\_ccw} / \texttt{rotate\_cw} apply roll (rotation about the local $Z$ axis), producing on-screen counter-clockwise / clockwise content rotation.

\paragraph{Discrete snapping.}
In discrete mode, after each rotation action the camera-to-world rotation matrix is decomposed into intrinsic $XYZ$ Euler angles, each angle is snapped to the nearest multiple of $s_r$, and the rotation matrix is recomposed.
This ensures that camera orientations remain on a regular grid, making action sequences exactly invertible.

\begin{table}[h]
\centering
\caption{Detailed action definitions. All rotations are about the camera center in local coordinates.}
\label{tab:action_space_detail}
\small
\setlength{\tabcolsep}{4pt}
\begin{tabular}{@{}llll@{}}
\toprule
Action & Type & Axis & Step size \\
\midrule
\texttt{move\_forward}  & Translation & Camera $+Z$ & $0.5$\,m \\
\texttt{move\_backward} & Translation & Camera $-Z$ & $0.5$\,m \\
\texttt{move\_left}     & Translation & Camera $-X$ & $0.5$\,m \\
\texttt{move\_right}    & Translation & Camera $+X$ & $0.5$\,m \\
\texttt{move\_up}       & Translation & Screen up ($-Y$) & $0.5$\,m \\
\texttt{move\_down}     & Translation & Screen down ($+Y$) & $0.5$\,m \\
\midrule
\texttt{turn\_left}     & Rotation & Yaw (local $Y$, $-$) & $30^\circ$ \\
\texttt{turn\_right}    & Rotation & Yaw (local $Y$, $+$) & $30^\circ$ \\
\texttt{look\_up}       & Rotation & Pitch (local $X$, $+$) & $30^\circ$ \\
\texttt{look\_down}     & Rotation & Pitch (local $X$, $-$) & $30^\circ$ \\
\texttt{rotate\_ccw}    & Rotation & Roll (local $Z$, $-$) & $30^\circ$ \\
\texttt{rotate\_cw}     & Rotation & Roll (local $Z$, $+$) & $30^\circ$ \\
\bottomrule
\end{tabular}
\end{table}

\subsection{Data Sampling and Filtering Pipeline}
\label{app:data_pipeline}

Algorithm~\ref{alg:data_pipeline} gives the pseudocode; Table~\ref{tab:pipeline_params} lists all hyperparameters.
Below we describe the four main stages.

\paragraph{Frame sampling.}
The temporal gap $\delta = f_\text{tgt} - f_\text{init}$ between initial and target frames is drawn from a mixture over three ranges: $\delta \in [50, 99]$ with weight $0.3$, $\delta \in [100, 300]$ with weight $0.5$, and the remaining frame indices uniformly with weight $0.2$.
The heavier weight on larger gaps means most pairs involve substantial viewpoint changes, though some nearby pairs are included as well.

\paragraph{Action planning.}
Given a sampled pair, we plan an action sequence from the initial to the target camera pose using a greedy, rotation-first strategy (Algorithm~\ref{alg:plan_actions}).
The six axes are processed in a fixed order (yaw, pitch, roll, forward, right, up); for each axis we try all step counts up to a maximum and pick the one that most reduces pose error, then commit those steps before moving on.
Because the planning is deterministic and single-pass, it consistently produces short sequences.
Pairs whose sequence length falls outside $[2, 10]$ are discarded.

\paragraph{Distractor generation.}
For the multiple-choice P2V and V2P tasks, we create $K{=}3$ distractors per pair by perturbing the ground-truth action sequence.
At $\lceil 0.3 \cdot \ell \rceil$ randomly chosen positions we apply one of three operations (replace with prob.\ $0.6$, remove $0.2$, insert $0.2$); replacements favor the same motion category with prob.\ $0.7$.
Each perturbed sequence is executed and rendered, and we reject any distractor whose mean pixel difference from every existing option is below $0.02$.

\paragraph{Scene-level filtering.}
In addition to the per-pair filters above (viewpoint identity, sequence length), we apply scene-level quality filtering based on the top-down reference view.
We first use a VLM to classify each scene's top-down view as \emph{good} (clear room layout visible from above, floor plan discernible) or \emph{bad} (mostly occluded by ceiling or other geometry, layout not discernible), using $12$ few-shot examples ($6$ good, $6$ bad).
The automated labels are then manually verified.
All view pairs from scenes classified as bad are removed from the dataset.
This filtering step removes scenes where the top-down view provides little useful spatial context, which would make the benchmark tasks ill-defined.

\begin{algorithm}[t]
\caption{\viewsuite{} data construction pipeline}
\label{alg:data_pipeline}
\small
\begin{algorithmic}[1]
\Require Point cloud $\mathcal{P}$; video frames with viewpoints $\{(f_i, P_i)\}_{i=1}^{N}$; delta distribution $\mathcal{D}$; length bounds $[\ell_\text{min}, \ell_\text{max}]$; num.\ distractors $K$; pixel threshold $\tau$
\State $V_\text{top} \gets \textsc{RenderTopDown}(\mathcal{P})$
\For{$n = 1, \dots, N_\text{pairs}$}
    \State Sample $\delta \sim \mathcal{D}$;\; sample $f_\text{init}$;\; $f_\text{tgt} \gets f_\text{init} + \delta$
    \State $P_\text{init}, P_\text{tgt} \gets$ viewpoints at $f_\text{init}, f_\text{tgt}$
    \If{$P_\text{init} \approx P_\text{tgt}$} \textbf{skip} \EndIf
    \State $\mathbf{a} \gets \textsc{PlanActions}(P_\text{init}, P_\text{tgt})$ \Comment{Alg.~\ref{alg:plan_actions}}
    \If{$|\mathbf{a}| \notin [\ell_\text{min},\, \ell_\text{max}]$} \textbf{skip} \EndIf
    \State $P_\text{tgt}^{*} \gets \textsc{Execute}(P_\text{init}, \mathbf{a})$ \Comment{Snap to discrete grid}
    \State $V_\text{init} \gets \textsc{Render}(\mathcal{P}, P_\text{init})$;\; $V_\text{tgt} \gets \textsc{Render}(\mathcal{P}, P_\text{tgt}^{*})$
    \State $\mathcal{O} \gets \{(\mathbf{a},\, V_\text{tgt})\}$ \Comment{Options: GT first}
    \For{$k = 1, \dots, K$} \Comment{Generate distractors}
        \Repeat
            \State $\hat{\mathbf{a}} \gets \textsc{Perturb}(\mathbf{a})$ \Comment{Replace / remove / insert ops}
            \State $\hat{V} \gets \textsc{Render}\bigl(\mathcal{P},\, \textsc{Execute}(P_\text{init}, \hat{\mathbf{a}})\bigr)$
        \Until{$\forall\, (\_,V') \in \mathcal{O}:\; \textsc{PixDiff}(\hat{V}, V') > \tau$}
        \State $\mathcal{O} \gets \mathcal{O} \cup \{(\hat{\mathbf{a}},\, \hat{V})\}$
    \EndFor
    \State Emit P2V, V2P, IVP instances from $(V_\text{init}, V_\text{top}, \mathcal{O})$
\EndFor
\end{algorithmic}
\end{algorithm}

\begin{algorithm}[t]
\caption{Greedy rotation-first action planning}
\label{alg:plan_actions}
\small
\begin{algorithmic}[1]
\Require Initial pose $P_\text{init}$; target pose $P_\text{tgt}$; axis order $\mathcal{A} = [\text{yaw, pitch, roll, fwd, right, up}]$; max steps per axis $k_\text{max}$
\Ensure Action sequence $\mathbf{a}$ such that $\textsc{Execute}(P_\text{init}, \mathbf{a}) \approx P_\text{tgt}$
\State $P_\text{cur} \gets P_\text{init}$;\; $\mathbf{a} \gets ()$
\For{each axis $(a^{+}, a^{-}) \in \mathcal{A}$}
    \State $e_0 \gets \textsc{PoseError}(P_\text{cur}, P_\text{tgt})$
    \State $k^{*} \gets \displaystyle\argmin_{k \in [-k_\text{max},\, k_\text{max}]} \textsc{PoseError}\bigl(\textsc{Execute}(P_\text{cur}, k),\; P_\text{tgt}\bigr) + 0.01|k|$
    \If{$\textsc{PoseError}(\textsc{Execute}(P_\text{cur}, k^{*}), P_\text{tgt}) < e_0$}
        \State Append $|k^{*}|$ copies of $(a^{+}$ if $k^{*} > 0$ else $a^{-})$ to $\mathbf{a}$
        \State $P_\text{cur} \gets \textsc{Execute}(P_\text{cur}, k^{*})$
    \EndIf
\EndFor
\State \Return $\mathbf{a}$
\end{algorithmic}
\end{algorithm}

\begin{table}[t]
\centering
\caption{Data pipeline hyperparameters.}
\label{tab:pipeline_params}
\small
\begin{tabular}{@{}lr@{}}
\toprule
Hyperparameter & Value \\
\midrule
\textit{Frame sampling} \\
\quad $\delta \in [50, 99]$ weight / $[100, 300]$ weight / complement & 0.3 / 0.5 / 0.2 \\
\quad Frame span (fraction of video) & $[0,\, 1]$ \\
\midrule
\textit{Action planning \& filtering} \\
\quad Axis order & Rot-first \\
\quad Sequence length bounds $[\ell_\text{min}, \ell_\text{max}]$ & $[2,\, 10]$ \\
\quad Max steps per axis (rotation / translation) & 12 / 10 \\
\midrule
\textit{Distractor generation} \\
\quad Num.\ distractors $K$ & 3 \\
\quad Perturb ratio $\lceil r \cdot \ell \rceil$ & $r{=}0.3$ \\
\quad Op probs (replace / remove / insert) & 0.6 / 0.2 / 0.2 \\
\quad Same-category replacement prob & 0.7 \\
\quad Pixel uniqueness threshold $\tau$ & 0.02 \\
\quad Max attempts per distractor & 20 \\
\midrule
\textit{Limits} \\
\quad Sampling attempts per pair & 20 \\
\quad Timeout per pair & 30\,s \\
\bottomrule
\end{tabular}
\end{table}

\subsection{Success Threshold Calibration}
\label{app:threshold_calibration}

We calibrate the threshold multipliers $\beta_t$ and $\beta_r$ in the success criterion of Section~\ref{sec:data} via a small human alignment study.
For each rollout we render the agent's submitted answer viewpoint with the same renderer and intrinsics used at evaluation time, and present it side by side with the ground-truth target view to expert annotators who judge whether the two views depict the same place (\emph{match}) or not.
We then sweep $(\beta_t, \beta_r)$ over translation thresholds $\{0.25,\, 0.5,\, 0.75,\, 1.0\}$\,m and rotation thresholds $\{30^\circ,\, 60^\circ,\, 90^\circ\}$, treating the threshold-based success indicator at each setting as a binary classifier of the human label.
Table~\ref{tab:threshold_calibration} reports precision, recall, $F_1$, and accuracy across the resulting $4\times 3$ grid.
The combination $0.5$\,m and $30^\circ$, equivalent to $(\beta_t, \beta_r)=(1, 1)$ given the discrete step sizes $s_t = 0.5$\,m and $s_r = 30^\circ$, achieves the highest $F_1$ ($0.915$) and accuracy ($0.920$) and is therefore adopted as the default success criterion throughout the paper.
Loosening either threshold further keeps recall essentially saturated but degrades precision sharply: with $\beta_t = 2$ ($1$\,m), precision drops to $0.72$ at $30^\circ$ and below $0.6$ at $60^\circ$, indicating that the human annotators consider many such viewpoints visibly different despite their proximity in pose space.

\begin{table}[h]
\centering
\caption{Success threshold calibration on IVP rollouts. Each row evaluates the threshold-based success indicator at one $(\text{position},\, \text{rotation})$ threshold pair against the human label. The $0.5$\,m / $30^\circ$ setting achieves the best agreement ($F_1{=}0.915$) and is adopted throughout the paper.}
\label{tab:threshold_calibration}
\small
\setlength{\tabcolsep}{6pt}
\begin{tabular}{@{}rrcccc@{}}
\toprule
Position thr. & Rotation thr. & Precision & Recall & F1 & Accuracy \\
\midrule
$0.25$\,m & $30^\circ$ & \textbf{1.000} & 0.600 & 0.750 & 0.820 \\
$0.25$\,m & $60^\circ$ & 0.931 & 0.600 & 0.730 & 0.800 \\
$0.25$\,m & $90^\circ$ & 0.931 & 0.600 & 0.730 & 0.800 \\
\midrule
$0.50$\,m & $30^\circ$ & 0.878 & 0.956 & \textbf{0.915} & \textbf{0.920} \\
$0.50$\,m & $60^\circ$ & 0.843 & 0.956 & 0.896 & 0.900 \\
$0.50$\,m & $90^\circ$ & 0.843 & 0.956 & 0.896 & 0.900 \\
\midrule
$0.75$\,m & $30^\circ$ & 0.811 & 0.956 & 0.878 & 0.880 \\
$0.75$\,m & $60^\circ$ & 0.694 & 0.956 & 0.804 & 0.790 \\
$0.75$\,m & $90^\circ$ & 0.683 & 0.956 & 0.796 & 0.780 \\
\midrule
$1.00$\,m & $30^\circ$ & 0.721 & \textbf{0.978} & 0.830 & 0.820 \\
$1.00$\,m & $60^\circ$ & 0.611 & \textbf{0.978} & 0.752 & 0.710 \\
$1.00$\,m & $90^\circ$ & 0.587 & \textbf{0.978} & 0.733 & 0.680 \\
\bottomrule
\end{tabular}
\end{table}

\subsection{View Distance Distribution}
\label{app:view_distance_dist}

\begin{figure}[h]
  \centering
  \includegraphics[width=0.55\columnwidth]{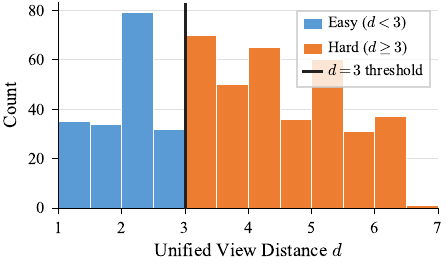}
  \caption{Distribution of unified view distance $d$ across $530$ test pairs. The threshold $d = 3$ separates \textsc{Short} ($185$ pairs) and \textsc{Long} ($345$ pairs) subsets.}
  \label{fig:view_distance_dist}
\end{figure}

Figure~\ref{fig:view_distance_dist} shows the empirical distribution of the unified view distance $d$ (Section~\ref{sec:viewsuite}) across the $530$ test pairs.
The distribution spans roughly $1.4$ to $6.8$ with mean $3.7$, indicating that most pairs require several atomic actions to traverse rather than trivial single-step adjustments.
We split the test set at $d = 3$ into a \textsc{Short} subset ($185$ pairs, $d < 3$) and a \textsc{Long} subset ($345$ pairs, $d \geq 3$); these subsets are used in the difficulty-stratified analysis of Section~\ref{sec:evaluation}.
The threshold $d = 3$ corresponds to about three atomic-action units of separation between initial and target viewpoints, which empirically produces a clean visual divide between trajectories that can typically be solved with a few correct actions and those that demand sustained planning.

\subsection{Task Examples}
\label{app:task_examples}

We show one example for each of the three tasks, including the system prompt and user prompt given to the model. Images are rendered from ScanNet point clouds. Placeholder \texttt{[image]} tokens mark where images are inserted in the multimodal input.

\paragraph{Path-to-View (P2V).}
Figure~\ref{fig:p2v_example} shows a P2V instance. The full prompt is given below.

\begin{quote}
\small
\textbf{System:} You are a spatial reasoning agent. You are given a question and a set of images. You need to answer the question based on the images. You can think first, which is optional, then answer, respond in this format: \texttt{<think>...</think><action>answer(x)</action>} where x is A, B, C, or D.

\textbf{User:} Given the initial view \texttt{[image]} and a top-down reference \texttt{[image]}, after you execute the following action sequence (translation step = 0.5\,m; rotation step = 30.0 degrees per step): [\texttt{turn\_right}, \texttt{turn\_right}, \texttt{turn\_right}, \texttt{turn\_right}, \texttt{turn\_right}], which of the following images corresponds to the result? A.\,\texttt{[image]} B.\,\texttt{[image]} C.\,\texttt{[image]} D.\,\texttt{[image]}
\end{quote}

GPT-5.4 Pro selects option C (incorrect), illustrating that even strong models struggle with large cumulative rotations.

\begin{figure}[h]
\centering
\begin{tabular}{cc}
\includegraphics[width=0.22\textwidth]{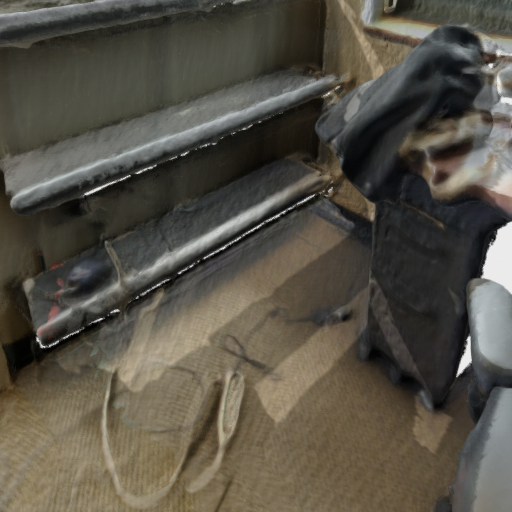} &
\includegraphics[width=0.22\textwidth]{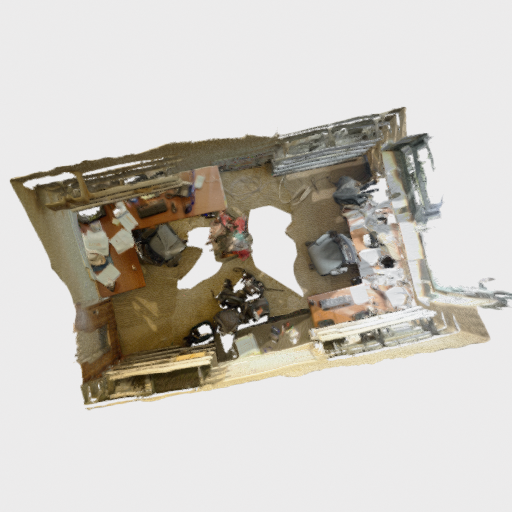} \\
{\small (a) Initial view} & {\small (b) Top-down view} \\[6pt]
\includegraphics[width=0.22\textwidth]{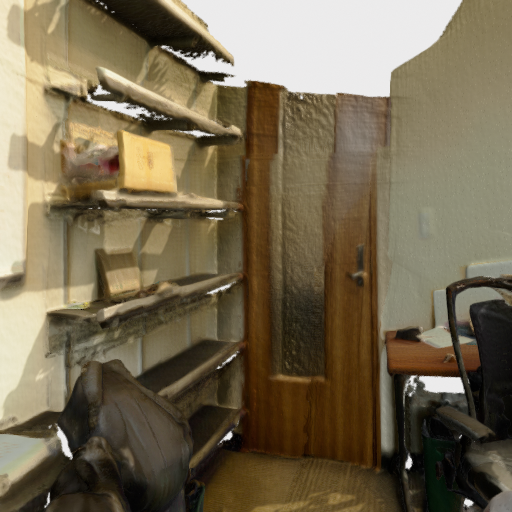} &
\includegraphics[width=0.22\textwidth]{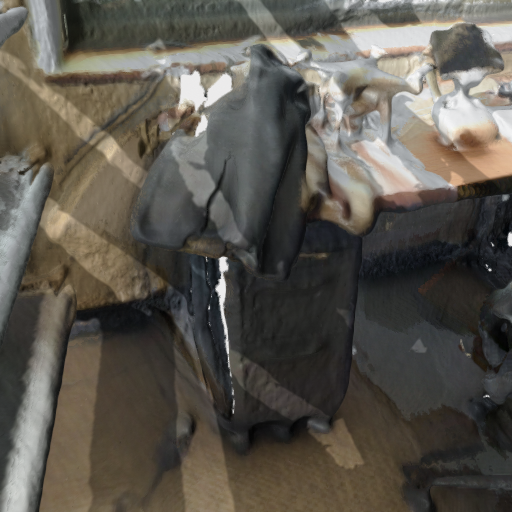} \\
{\small (c) Option A} & {\small (d) Option B} \\[6pt]
\includegraphics[width=0.22\textwidth]{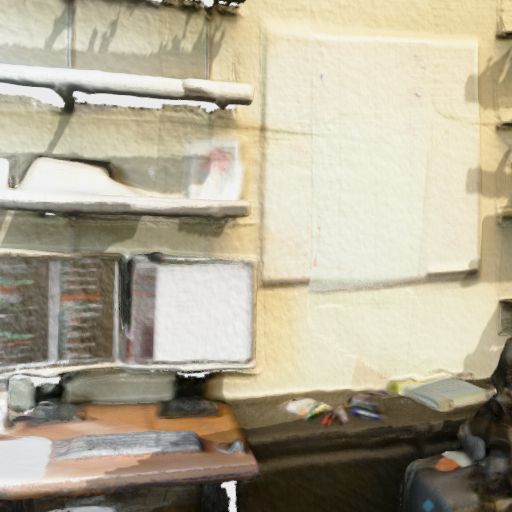} &
\includegraphics[width=0.22\textwidth]{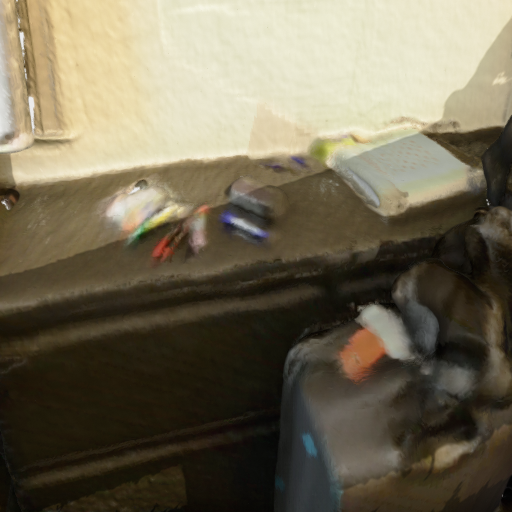} \\
{\small (e) Option C} & {\small (f) Option D} \\
\end{tabular}
\caption{Path-to-View (P2V) example: given the initial view (a), the top-down view (b), and an action sequence, the model picks the resulting view among options (c)--(f). Action: [\texttt{turn\_right} $\times$ 5]. GPT-5.4 Pro selects C (incorrect).}
\label{fig:p2v_example}
\end{figure}

\paragraph{View-to-Path (V2P).}
Figure~\ref{fig:v2p_example} shows a V2P instance. The full prompt is given below.

\begin{quote}
\small
\textbf{System:} You are a spatial reasoning agent. You are given a question and a set of images. You need to answer the question based on the images. You can think first, which is optional, then answer, respond in this format: \texttt{<think>...</think><action>answer(x)</action>} where x is A, B, C, or D.

\textbf{User:} Given the initial view \texttt{[image]} and a top-down reference \texttt{[image]}, which action sequence will reach the target view \texttt{[image]}? (Action semantics: translation step = 0.5\,m; rotation step = 30.0 degrees per step.)\\
A.\,[\texttt{look\_up}, \texttt{move\_forward}, \texttt{move\_left}]\\
B.\,[\texttt{turn\_left} $\times$ 5, \texttt{move\_left}]\\
C.\,[\texttt{turn\_right} $\times$ 2, \texttt{move\_forward}, \texttt{move\_left} $\times$ 5, \texttt{move\_up}]\\
D.\,[\texttt{turn\_left} $\times$ 2]
\end{quote}

GPT-5.4 Pro correctly selects B, reasoning that the target view is behind the initial direction with reversed wall orientation, consistent with a large left rotation plus a lateral shift.

\begin{figure}[h]
\centering
\begin{tabular}{ccc}
\includegraphics[width=0.22\textwidth]{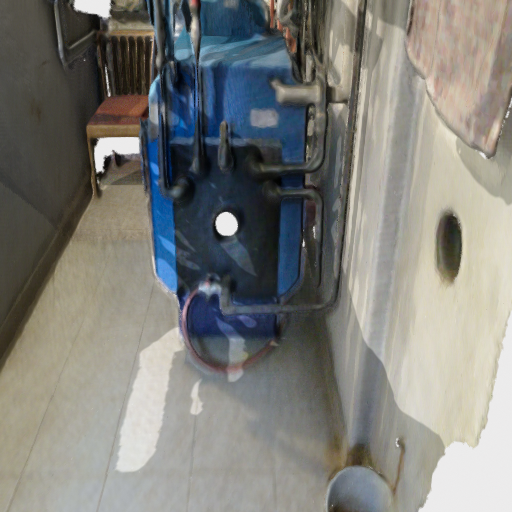} &
\includegraphics[width=0.22\textwidth]{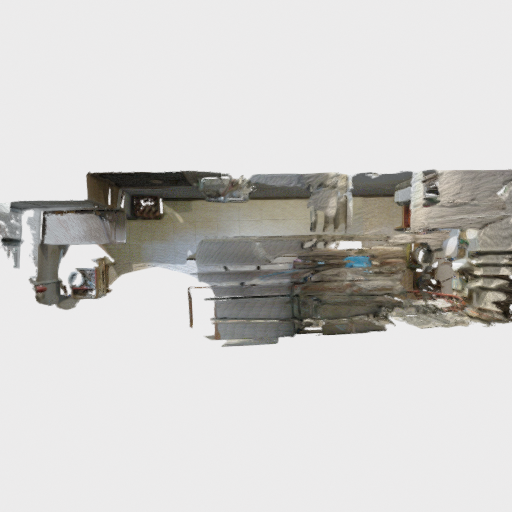} &
\includegraphics[width=0.22\textwidth]{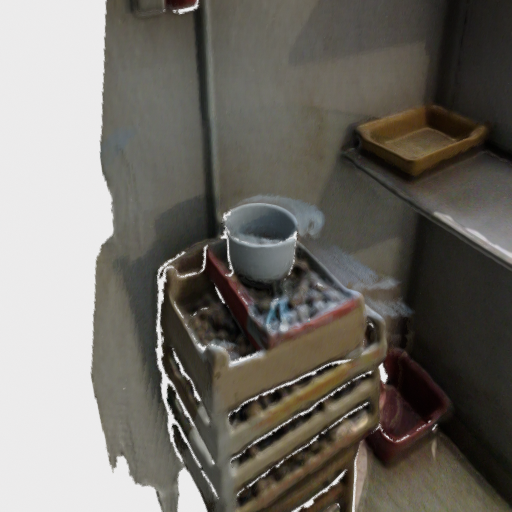} \\
{\small (a) Initial view} & {\small (b) Top-down view} & {\small (c) Target view} \\
\end{tabular}
\caption{View-to-Path (V2P) example: given the initial (a), top-down (b), and target (c) views, the model picks which of four action sequences connects the initial view to the target view. GPT-5.4 Pro selects B (correct): [\texttt{turn\_left} $\times$ 5, \texttt{move\_left}].}
\label{fig:v2p_example}
\end{figure}

\begin{figure}[h]
\centering
\begin{tabular}{ccc}
\includegraphics[width=0.22\textwidth]{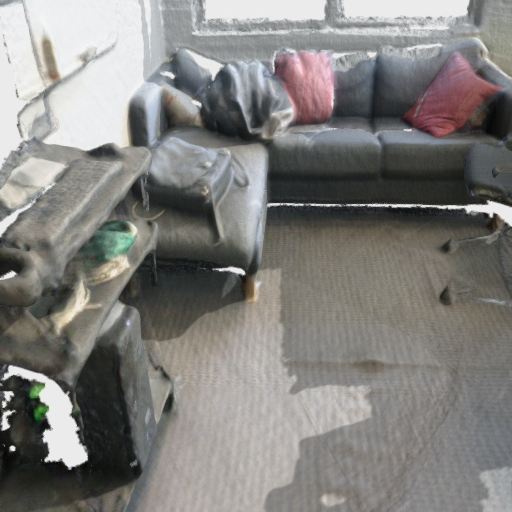} &
\includegraphics[width=0.22\textwidth]{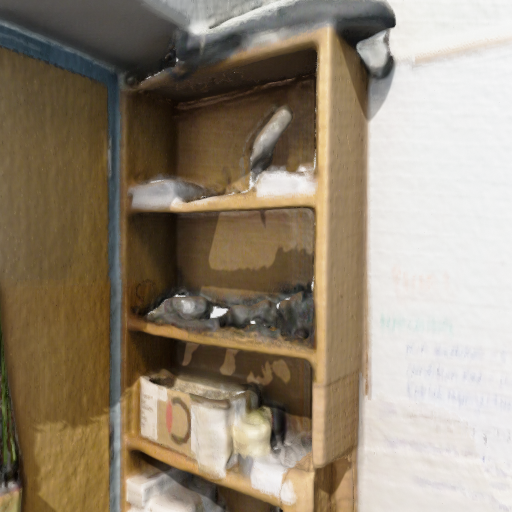} &
\includegraphics[width=0.22\textwidth]{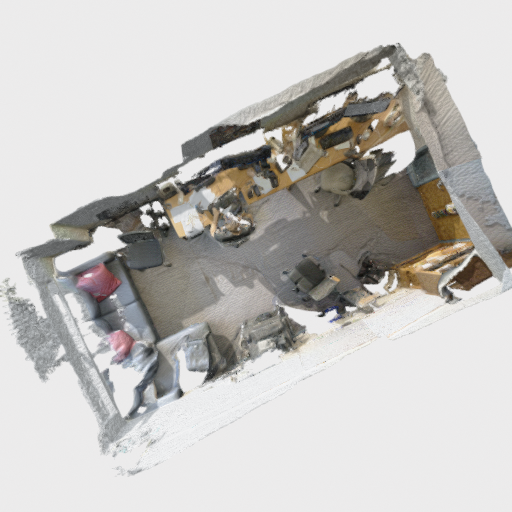} \\
{\small (a) Target view} & {\small (b) Initial view} & {\small (c) Top-down view} \\[6pt]
\includegraphics[width=0.22\textwidth]{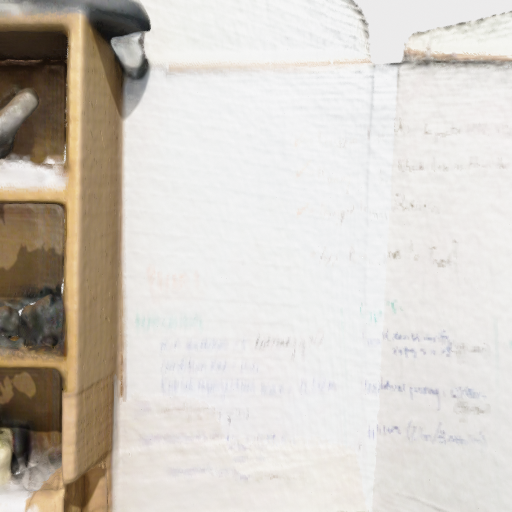} &
\includegraphics[width=0.22\textwidth]{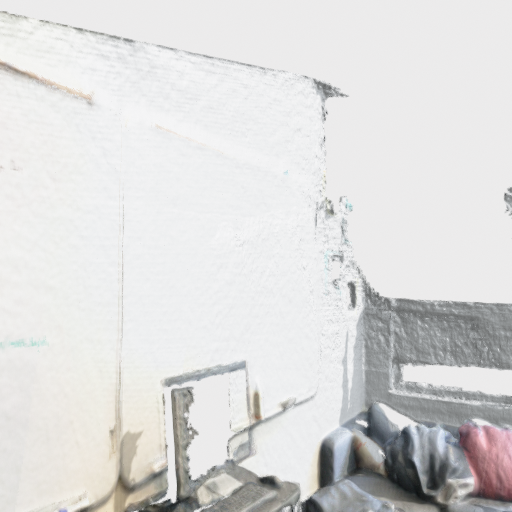} &
\includegraphics[width=0.22\textwidth]{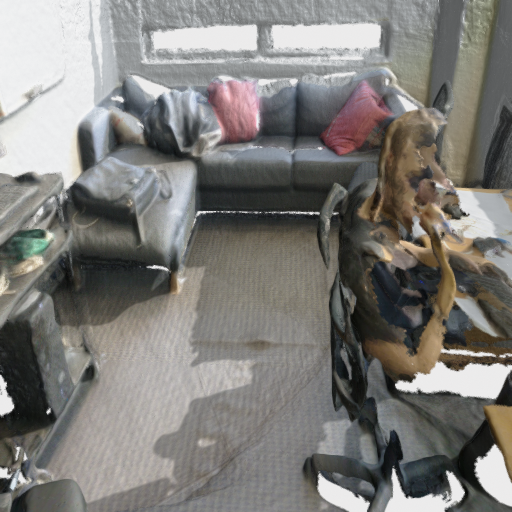} \\
{\small (d) After step 1} & {\small (e) After step 2} & {\small (f) After step 3} \\[6pt]
\includegraphics[width=0.22\textwidth]{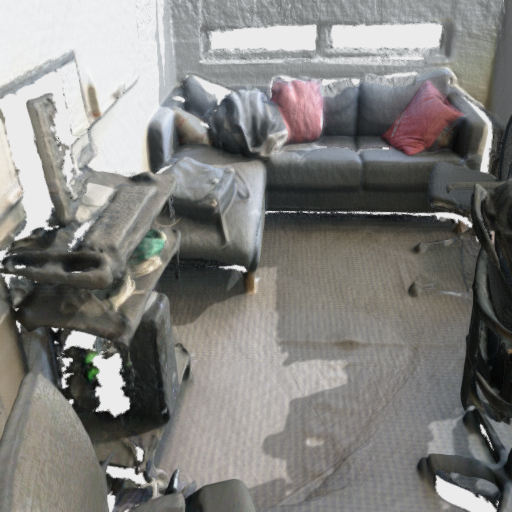} &
\includegraphics[width=0.22\textwidth]{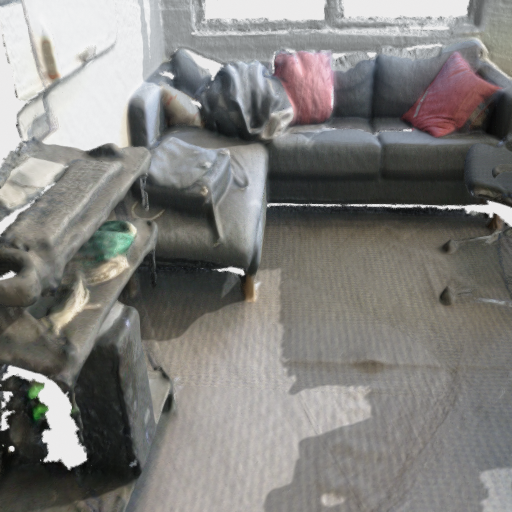} & \\
{\small (g) After step 4} & {\small (h) After step 5 (final)} & \\
\end{tabular}
\caption{Interactive View Planning (IVP) example. Our trained Qwen2.5-VL-7B plans view changes from the initial view (b) to localize the target view (a) in $6$ turns. Final pose error: $0.061$\,m / $0^\circ$. Success.}
\label{fig:tvn_example}
\end{figure}

\paragraph{Interactive View Planning (IVP).}
Figure~\ref{fig:tvn_example} shows an IVP instance solved by our trained Qwen2.5-VL-7B. The system prompt is given below (abridged; full action list omitted for space).

\begin{quote}
\small
\textbf{System:} You are solving an interactive view-planning camera pose estimation task.

\textsc{Goal}: Predict the target view absolute camera pose (camera-to-world, c2w) as a 6-DoF vector: [tx, ty, tz, rx, ry, rz]. You may explore the 3D scene using camera-control actions, then submit a final answer. Your predicted pose should be as close as possible to the target pose.

\textsc{Turn limit}: You must complete the task within 10 turns, including the final answer.

\textsc{Output format}:\\
\texttt{<think>...</think><action>action\_1|action\_2|...</action>}. The final response must contain exactly one \texttt{answer(tx, ty, tz, rx, ry, rz)}.

\textbf{User:} You're in scene \texttt{scene0474\_00}. Please study the target view \texttt{[image]}, the initial view \texttt{[image]}, and the top-down view \texttt{[image]}. You start from the initial view. Move toward the target view using actions. Initial view camera 6-DoF: [tx=4.07, ty=3.28, tz=1.66, rx=$-90^\circ$, ry=$0^\circ$, rz=$-120^\circ$]. Success thresholds: position error $\leq$ 0.5\,m, rotation error $\leq$ $30^\circ$.
\end{quote}

Over $6$ turns, the agent executes: \texttt{turn\_right} (step 1), \texttt{turn\_right} $\times$ 2 (step 2), \texttt{turn\_right}, \texttt{look\_down} (step 3), \texttt{move\_left} (step 4), \texttt{move\_forward} (step 5), then submits a pose estimate (step 6). The final pose error is $0.061$\,m position and $0^\circ$ rotation, well within the success threshold.

\section{Extended Evaluation Results}
\label{app:evaluation}

\subsection{Evaluation-Protocol Ablations on IVP}
\label{app:protocol_ablations}

The IVP results in Table~\ref{tab:main_results} and Table~\ref{tab:training_results} follow our default evaluation protocol: at each turn the agent's rotation actions are rounded (\emph{snapped}) to integer multiples of the discrete step size $s_r{=}30^\circ$, and an episode succeeds only if the agent issues an explicit \texttt{submit} action while its pose lies within the unified-distance threshold of the target view. To check that our findings do not hinge on these two protocol details, we re-evaluate our fully-trained models alongside two strong proprietary baselines (Gemini 3.1 Pro and GPT-5.4) under two relaxations:
\begin{itemize}[leftmargin=*,nosep]
\item \textbf{No-Snap.} Per-step rotations are no longer rounded to step-size multiples; the raw rotation magnitudes emitted by the agent are executed as-is. This isolates whether the planning gains depend on the discrete action grid.
\item \textbf{No-Submit.} The agent does not need to explicitly submit. An episode is counted as successful at the first turn its pose enters the unified-distance threshold of the target, similar to a pure ``reach the goal'' criterion used in standard navigation benchmarks.
\end{itemize}

Table~\ref{tab:protocol_ablations} compares the default protocol against the two ablations on the Short / Long splits defined in Section~\ref{sec:viewsuite}. The ordering between models is preserved under all three protocols: our trained models continue to outperform Gemini 3.1 Pro and GPT-5.4 by a wide margin (e.g., $19.6$ vs.\ $15.7$ on No-Snap, $60.2$ vs.\ $31.5$ on No-Submit). Relative to the default protocol, No-Snap \emph{lowers} overall success for every model: without rounding, per-step rotation residuals accumulate over the $10$-turn horizon, and the agent drifts off the on-grid pose distribution from which target views are drawn. No-Submit instead \emph{raises} success, since the criterion no longer requires the agent to commit to a final answer and credits any turn at which its pose enters the unified-distance threshold. Across all three protocols the Qwen2.5-VL-7B backbone outperforms Qwen3-VL-8B ($47.8$ vs.\ $32.5$ on Default, $19.6$ vs.\ $18.5$ on No-Snap, $60.2$ vs.\ $48.3$ on No-Submit), reinforcing that our framework's gains transfer to both backbones, with Qwen2.5-VL-7B being the stronger starting point on this benchmark.

\paragraph{Coverage note for GPT-5.4 Pro.} GPT-5.4 Pro declines a subset of IVP episodes under its content policy, returning a refusal in place of an action. Of the $530$ IVP test instances, $23$ are refused in this way and produce no valid rollout; we therefore report GPT-5.4 Pro's IVP success rate over the remaining $507$ valid episodes ($101/507 = 19.9\%$). All other models complete all $530$ instances. This affects IVP only: GPT-5.4 Pro completes the full $530$ test pairs on both P2V and V2P.

\begin{table}[h]
\centering
\caption{\textbf{The method ranking is protocol-independent.} IVP success rates (\%) under the default protocol and two evaluation-protocol ablations. \textit{Default}: per-step rotations are snapped to integer multiples of $s_r{=}30^\circ$, and success requires an explicit \texttt{submit} within the unified-distance threshold; numbers are reproduced from Table~\ref{tab:main_results} and Table~\ref{tab:training_results}. \textit{No-Snap}: rotations are not snapped; raw rotation magnitudes are executed as-is. \textit{No-Submit}: no explicit \texttt{submit} required; an episode is counted as successful as soon as its pose falls within the unified-distance threshold of the target view. ``Ours'' denotes our fully-trained models (Section~\ref{sec:main_results_training}).}
\label{tab:protocol_ablations}
\small
\setlength{\tabcolsep}{3.5pt}
\begin{tabular}{@{}l ccc ccc ccc@{}}
\toprule
& \multicolumn{3}{c}{Default} & \multicolumn{3}{c}{No-Snap} & \multicolumn{3}{c}{No-Submit} \\
\cmidrule(lr){2-4} \cmidrule(lr){5-7} \cmidrule(lr){8-10}
Method & Short & Long & All & Short & Long & All & Short & Long & All \\
\midrule
Gemini 3.1 Pro & 28.6 & 17.4 & 21.3 & 27.0 & 9.6 & 15.7 & 49.7 & 21.7 & 31.5 \\
GPT-5.4 & 33.5 & 7.5 & 16.6 & 27.6 & 5.2 & 13.0 & 57.3 & 17.4 & 31.3 \\
\midrule
Ours (Qwen2.5-VL-7B) & 67.2 & 36.9 & 47.8 & 38.9 & 9.3 & 19.6 & 79.5 & 49.9 & 60.2 \\
Ours (Qwen3-VL-8B) & 56.8 & 19.4 & 32.5 & 41.6 & 6.1 & 18.5 & 80.5 & 31.0 & 48.3 \\
\bottomrule
\end{tabular}
\end{table}

\subsection{Sample-Level Factor Analysis}
\label{app:factor_analysis}

We compute Spearman $\rho$ between $12$ sample-level factors and per-model binary success across all three tasks (Figure~\ref{fig:factor_heatmap}; factor definitions in Appendix~\ref{app:factor_definitions}).
Factors span geometric distance, visual overlap (from pointcloud coverage), and directional geometry.

Across all tasks, distance factors show consistent negative correlations: farther view pairs are harder.
For P2V and V2P, orientation agreement is the strongest positive predictor ($\rho \approx +0.19$--$+0.30$), confirming that same-facing camera pairs are easier to reason about.
For IVP, position distance dominates ($\rho$ up to $-0.42$ for GPT-5.4 Pro), consistent with the position-bottleneck finding in Section~\ref{sec:eval_analysis}.
Visual overlap factors show mild positive correlations for P2V/V2P, indicating that shared visual content helps single-turn prediction.
One notable outlier is Grok 4.20 Beta on V2P, which shows near-zero or slightly positive correlations with distance factors, suggesting a qualitatively different (and possibly less spatially grounded) reasoning strategy.

\begin{figure*}[h]
\centering
\includegraphics[width=\textwidth]{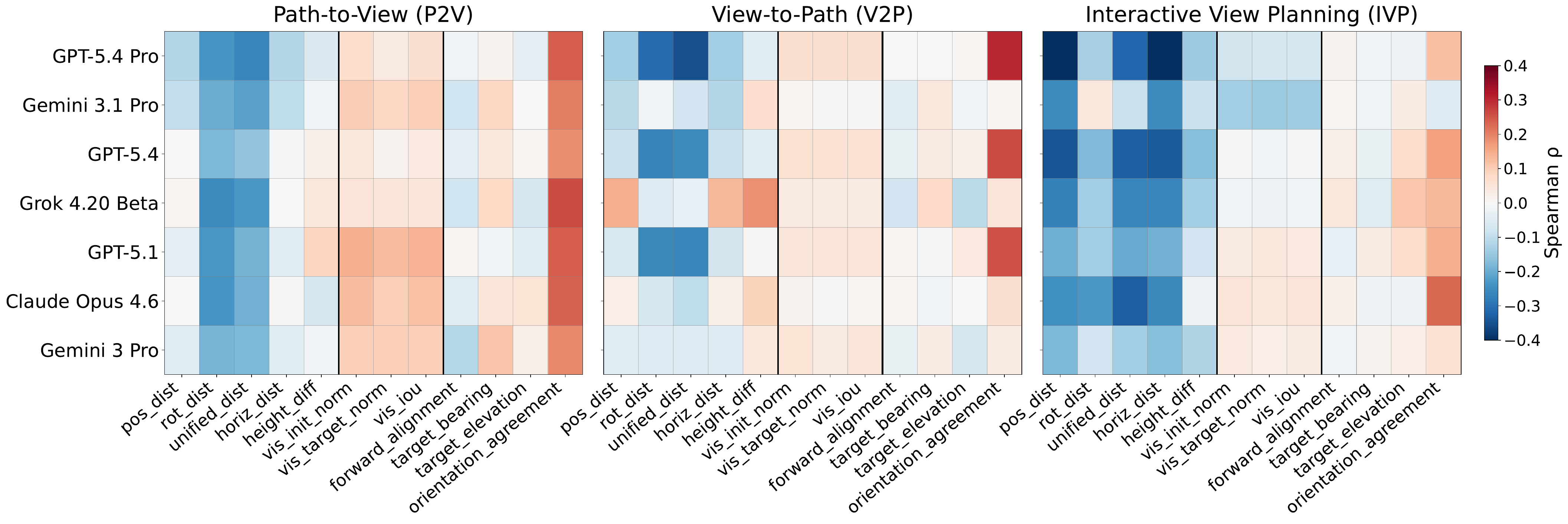}
\caption{Spearman $\rho$ between sample-level factors (rows) and per-model success (columns), grouped into geometric distance, visual overlap, and directional geometry. Position distance is the strongest predictor of IVP success; orientation agreement is the strongest for P2V/V2P.}
\label{fig:factor_heatmap}
\end{figure*}

\subsection{Success Factor Definitions}
\label{app:factor_definitions}

We define the $12$ sample-level factors used in Figure~\ref{fig:factor_heatmap}. All factors are computed from the initial and target camera-to-world extrinsics ($4{\times}4$ matrices) shared across the $530$ test view pairs. We write position $\mathbf{t} = C_{[:3,3]} \in \mathbb{R}^3$ (the translation column), rotation $R = C_{[:3,:3]} \in \mathbb{R}^{3\times3}$ (the rotation submatrix), and camera forward direction $\mathbf{f} = -R_{[:,2]} \in \mathbb{R}^3$ (negative z-axis of the camera frame, transformed to world coordinates).

\paragraph{Group A: Geometric Distance.}
\begin{itemize}[leftmargin=*,nosep,itemsep=3pt]
\item \texttt{pos\_dist} ($d_\text{pos}$): $\|\mathbf{t}_\text{init} - \mathbf{t}_\text{target}\|_2$ (meters). Euclidean distance between camera positions.
\item \texttt{rot\_dist} ($d_\text{rot}$): $\arccos\!\bigl(\text{clip}\bigl(\frac{\text{tr}(R_\text{init}^\top R_\text{target})-1}{2},\,-1,\,1\bigr)\bigr)$ (degrees). Geodesic angle between orientations.
\item \texttt{unified\_dist}: $\sqrt{(d_\text{pos}/s_t)^2 + (d_\text{rot}/s_r)^2}$ (steps), where $s_t = 0.5$\,m and $s_r = 30^\circ$ are the discrete step sizes in \viewsuite{}. Equivalent to the unified view distance $d$ defined in Section~\ref{sec:viewsuite}.
\item \texttt{horiz\_dist}: $\|\mathbf{t}_\text{init}^{xy} - \mathbf{t}_\text{target}^{xy}\|_2$ (meters). Horizontal distance, ignoring vertical displacement.
\item \texttt{height\_diff}: $|\mathbf{t}_\text{init}^z - \mathbf{t}_\text{target}^z|$ (meters). Absolute vertical difference.
\end{itemize}

\paragraph{Group B: Visual Overlap.}
Computed from GPU-rendered pointcloud coverage: for each viewpoint, we determine which mesh vertices are visible via depth rendering, yielding vertex sets $V_\text{init}$ and $V_\text{target}$.
\begin{itemize}[leftmargin=*,nosep,itemsep=3pt]
\item \texttt{vis\_init\_norm}: $|V_\text{init} \cap V_\text{target}| \,/\, |V_\text{init}|$. Fraction of init-visible vertices also visible from target.
\item \texttt{vis\_target\_norm}: $|V_\text{init} \cap V_\text{target}| \,/\, |V_\text{target}|$. Fraction of target-visible vertices already visible from init.
\item \texttt{vis\_iou}: $|V_\text{init} \cap V_\text{target}| \,/\, |V_\text{init} \cup V_\text{target}|$. Intersection-over-union of visible vertex sets.
\end{itemize}

\paragraph{Group C: Directional Geometry.}
Let $\hat{\mathbf{d}}$ denote the unit displacement vector from the initial to the target position.
\begin{itemize}[leftmargin=*,nosep,itemsep=3pt]
\item \texttt{forward\_alignment}: $\hat{\mathbf{f}}_\text{init} \cdot \hat{\mathbf{d}}$. Ranges from $+1$ (target ahead) to $-1$ (target behind).
\item \texttt{target\_bearing}: $\arccos(\text{clip}(\texttt{forward\_alignment},\,-1,\,1))$ (degrees). Angle between init forward direction and displacement to target.
\item \texttt{target\_elevation}: $\text{atan2}(\Delta z,\, \|\Delta_{xy}\|)$ (degrees). Vertical angle from init to target.
\item \texttt{orientation\_agreement}: $\hat{\mathbf{f}}_\text{init} \cdot \hat{\mathbf{f}}_\text{target}$. Cosine between camera forward directions. $+1$ = same facing, $-1$ = opposite.
\end{itemize}

\section{Iterative Training Implementation Details}
\label{app:ivet}

\subsection{Algorithm}
\label{app:algorithm}

Algorithm~\ref{alg:ivet} summarizes our iterative framework, alternating self-exploration with view graph distillation. Each iteration appends new trajectories to a persistent view graph, samples paths from it, and reformulates them via Eq.~\ref{eq:reformulation} into supervised view-planning demonstrations.

\begin{algorithm}[h]
\caption{Self-Exploration with View Graph Distillation}
\label{alg:ivet}
\begin{algorithmic}[1]
\Require initial policy $\pi_{\theta_0}$, environments $\mathcal{E}$, iterations $K$
\State $G_0 \gets \emptyset$ \Comment{empty view graph}
\For{$k = 0, 1, \ldots, K-1$}
    \State \textbf{Self-exploration stage:}
    \State \quad Run PPO updates of $\pi_{\theta_k}$ on $\mathcal{E}$ with reward Eq.~\ref{eq:reward}
    \State \quad Append trajectories: $G_{k+1} \gets G_k \cup \mathrm{traj}(\pi_{\theta_k})$
    \State \textbf{View graph distillation stage:}
    \State \quad Sample paths $\{P_i\} \subset G_{k+1}$
    \State \quad Reformulate: $\mathcal{D}_{k+1} \gets \{\mathcal{R}(P_i)\}$ via Eq.~\ref{eq:reformulation}
    \State \quad Fine-tune via SFT: $\theta_{k+1} \gets \arg\min_\theta \mathcal{L}_{\mathrm{SFT}}(\theta; \mathcal{D}_{k+1})$
\EndFor
\State \Return $\pi_{\theta_K}$
\end{algorithmic}
\end{algorithm}

\subsection{RL Hyperparameters}
\label{app:rl_hyperparams}

Table~\ref{tab:rl_hyperparams} lists the RL training hyperparameters used across all iterations of our framework and RL baselines.
All methods use the same PPO configuration unless otherwise noted.

\begin{table}[h]
\centering
\caption{RL training hyperparameters for our framework and baselines.}
\label{tab:rl_hyperparams}
\small
\begin{tabular}{@{}lr@{}}
\toprule
Hyperparameter & Value \\
\midrule
\textit{Algorithm} \\
\quad Advantage estimator & GAE \\
\midrule
\textit{Actor} \\
\quad Learning rate & $1 \times 10^{-6}$ \\
\quad Mini batch size & 128 \\
\quad Micro batch size per GPU & 2 \\
\quad FSDP param offload & True \\
\quad FSDP optimizer offload & True \\
\quad Gradient checkpointing & True \\
\midrule
\textit{Critic} \\
\quad Learning rate & $1 \times 10^{-5}$ \\
\quad Micro batch size per GPU & 2 \\
\quad FSDP param offload & True \\
\quad FSDP optimizer offload & True \\
\quad Critic warmup steps & 0 \\
\midrule
\textit{Rollout} \\
\quad Engine & SGLang (async) \\
\quad Max batched tokens & 32{,}768 \\
\quad GPU memory utilization & 0.6 \\
\quad Tensor parallel size & 1 \\
\midrule
\textit{Data} \\
\quad Max prompt length & 4{,}000 \\
\quad Max response length & 10{,}000 \\
\quad Train batch size & 128 \\
\midrule
\textit{Infrastructure} \\
\quad GPUs per node & 8 \\
\quad Nodes & 1 \\
\bottomrule
\end{tabular}
\end{table}

\paragraph{Iteration-specific overrides.}
Iterations 0--2 use $60$ RL training steps each for rapid bootstrapping.
The final iteration (iteration $3$) is trained to convergence.

\paragraph{Direct GRPO baseline.}
The Direct GRPO (filter) baseline uses identical infrastructure but replaces GAE with the GRPO advantage estimator~\citep{grpo}, sets $n{=}4$ rollouts per prompt for filtering, and trains for $1{,}000$ steps.

\subsection{SFT Hyperparameters}
\label{app:sft_hyperparams}

Table~\ref{tab:sft_hyperparams} lists the SFT hyperparameters used in our framework.

\begin{table}[h]
\centering
\caption{SFT training hyperparameters for our framework.}
\label{tab:sft_hyperparams}
\small
\begin{tabular}{@{}lr@{}}
\toprule
Hyperparameter & Value \\
\midrule
\textit{Training} \\
\quad Learning rate & $1 \times 10^{-5}$ \\
\quad Weight decay & 0.01 \\
\quad LR scheduler & Cosine \\
\quad Warmup ratio & 0.1 \\
\quad Per-device batch size & 2 \\
\quad Gradient accumulation steps & 2 \\
\quad Cutoff length & 16{,}384 \\
\quad Precision & BF16 \\
\quad Flash attention & FA2 \\
\quad Distributed strategy & DeepSpeed ZeRO-2 \\
\midrule
\textit{Epochs} \\
\quad Iterations 0--2 & 3 \\
\quad Iteration 3 (final) & 4 \\
\midrule
\textit{Model selection} \\
\quad Validation split & 20\% \\
\quad Eval strategy & Per epoch \\
\quad Best model metric & Eval loss \\
\bottomrule
\end{tabular}
\end{table}

\subsection{View Graph Construction}
\label{app:graph_construction}

During RL exploration, a background process runs concurrently with training and incrementally merges completed trajectories into the view graph.

\paragraph{Node and edge representation.}
Each node stores a 6-DoF viewpoint (position and rotation) and its rendered view at $512 \times 512$ resolution.
Each directed edge stores the sequence of camera actions taken between two viewpoints.
Before adding a node, we apply image quality filters: frames with void fraction $> 0.7$ (indicating the camera is looking outside the point cloud) or pixel standard deviation $< 10.0$ (indicating a near-uniform, uninformative view) are discarded.

\paragraph{Deduplication.}
Nodes are deduplicated by viewpoint similarity: a new node is merged with an existing node if their position distance is below $0.25$\,m \emph{and} rotation distance is below $15^\circ$.
When two nodes are merged, all edges incident to the new node are redirected to the existing node.
Edges are then deduplicated by (source, target, action sequence) identity.
This deduplication prevents the graph from growing unboundedly as the same regions of the scene are revisited across episodes.

\paragraph{Cross-iteration accumulation.}
The graph is persisted to disk and accumulated across all self-exploration iterations.
Later distillation stages sample from the full exploration history, not just the most recent iteration.
This means that spatial knowledge discovered in early iterations (when the policy is weak) remains available for training in later iterations, even if the improved policy explores different regions.

\begin{figure}[h]
\centering
\includegraphics[width=0.65\columnwidth]{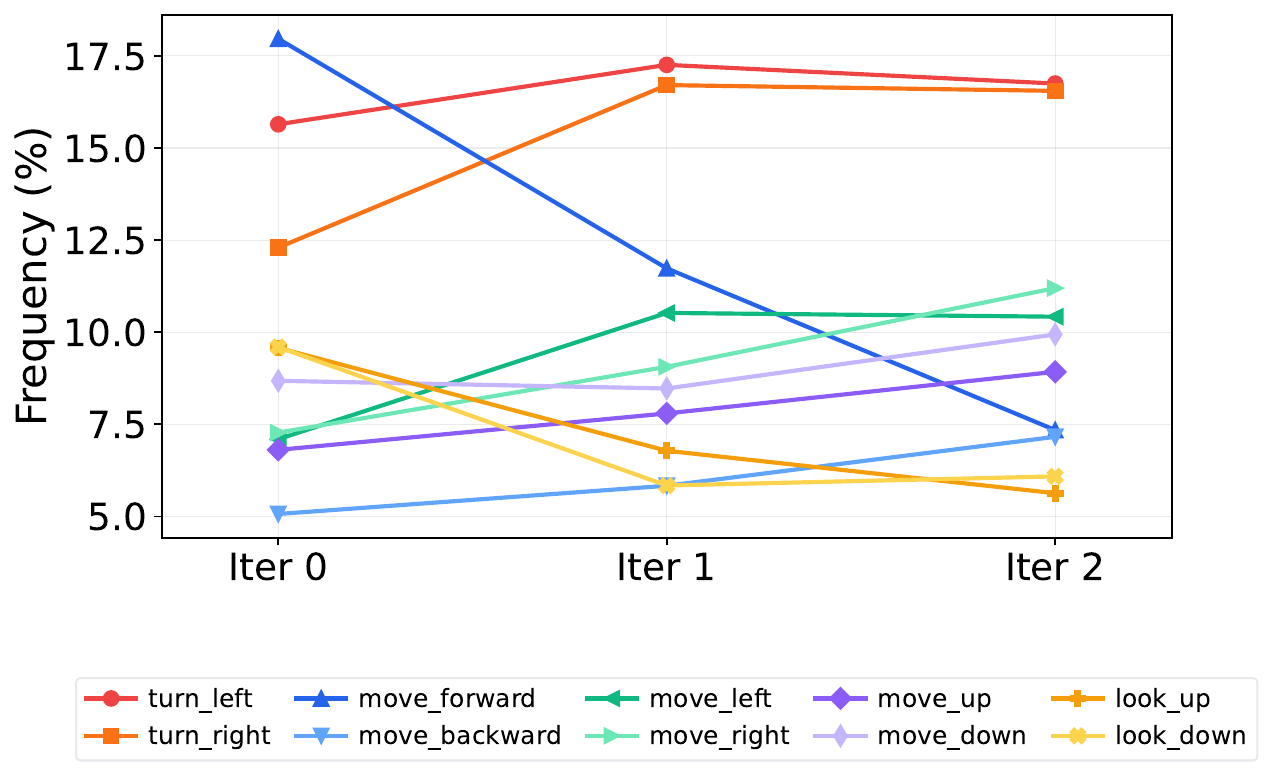}
\caption{Action frequency distribution of self-exploration across training iterations. The base policy (iteration~0) favors \texttt{move\_forward}; later iterations shift toward rotations and more balanced translations.}
\label{fig:action_dist}
\end{figure}

Table~\ref{tab:graph_growth} shows the graph growth across iterations.
The graph grows by an order of magnitude from iteration~0 to iteration~1 as the bootstrapped policy explores more effectively, then grows incrementally in iteration~2.
Figure~\ref{fig:action_dist} shows how the action distribution shifts across iterations.
In iteration~0, \texttt{move\_forward} dominates ($18.0\%$), reflecting the base policy's tendency to move straight ahead.
By iteration~2, rotations (\texttt{turn\_left}, \texttt{turn\_right}) become the most frequent actions (${\sim}33\%$ combined), and translations become more balanced across all six directions, indicating that the trained policy has learned more diverse exploration strategies.

\begin{table}[h]
\centering
\caption{View graph growth across iterations (iterations 0--2; the final iteration uses RL only without graph construction). The graph grows by an order of magnitude from iteration~0 to iteration~1, then incrementally.}
\label{tab:graph_growth}
\small
\begin{tabular}{@{}lrrrrr@{}}
\toprule
Iteration & Scenes & Nodes & Edges & Avg Nodes/Scene & Avg Actions/Edge \\
\midrule
0 & 186 & 4{,}067 & 2{,}875 & 21.9 & 1.6 \\
1 & 193 & 61{,}862 & 62{,}445 & 320.5 & 1.5 \\
2 & 193 & 66{,}492 & 65{,}577 & 344.5 & 1.7 \\
\bottomrule
\end{tabular}
\end{table}

\subsection{Task Reformulation Details}
\label{app:task_reformulation}

We generate three supervision types from paths sampled from the view graph.
Table~\ref{tab:task_reformulation_params} summarizes the sampling parameters for each task type.

\begin{table}[h]
\centering
\caption{Task reformulation sampling parameters.}
\label{tab:task_reformulation_params}
\small
\begin{tabular}{@{}lccc@{}}
\toprule
Task Type & Path Length & Samples/Scene & Balanced \\
\midrule
Multi-turn view planning & 3--5 & 20 & No \\
View-difference estimation & 2--5 & 15 & Yes \\
View-difference MCQ & 2--5 & 15 & Yes \\
\bottomrule
\end{tabular}
\end{table}

\paragraph{Multi-turn view planning (primary task).}
For a sampled path of length $\ell$ ($3 \leq \ell \leq 5$ edges), the end node is designated as the target view and the start node as the initial view.
The intermediate nodes provide turn-by-turn observations.
The model is trained to predict the correct camera action at each turn, given the current view, the target view, and the planning history.
We oversample each path $10$ times with different random seeds to increase diversity.
This task directly trains the IVP capability.

\paragraph{View-difference estimation.}
Given two views sampled from nodes at path distance $\ell$ ($2 \leq \ell \leq 5$), the model predicts the unified view distance between them.
This auxiliary task encourages the model to develop a sense of spatial distance between views, complementing the view-planning task.
Balanced sampling ensures equal representation across path lengths.

\paragraph{View-difference MCQ.}
Same setup as view-difference estimation, but presented as a multiple-choice question with four options.
This provides an alternative answer format, preventing the model from overfitting to a single task format during SFT.

\paragraph{Additional reformulations (not used in main experiments).}
The view graph is a task-agnostic representation, and the three tasks above are only the subset we use for training.
The same graph naturally admits further reformulations.
\emph{Inverse dynamics}: given two views sampled as graph nodes, the model predicts the action sequence labeling the connecting edges.
\emph{Forward dynamics}: given an initial view and an action sequence, the model selects the resulting view from several candidate images.
We include these to show that distilling structured spatial knowledge from the graph is not tied to goal-conditioned relabeling; studying their effect on training is left to future work.

\subsection{Training and Validation Environments}
\label{app:train_val_envs}

\paragraph{Training.}
We use the $3{,}419$ \viewsuite{} training view pairs for RL, each defining one IVP environment (a scene with an initial and a target view).
Each episode runs for up to 10 turns at $512 \times 512$ image resolution, rendered via Open3d.

\paragraph{Validation.}
The validation set consists of $378$ instances each for the P2V and V2P, and $100$ instances for the IVP task.

\section{Extended Analysis}
\label{app:analysis}

\subsection{Point Cloud Coverage: Full Model Comparison}
\label{app:coverage_all}

\begin{figure*}[h]
\centering
\includegraphics[width=\textwidth]{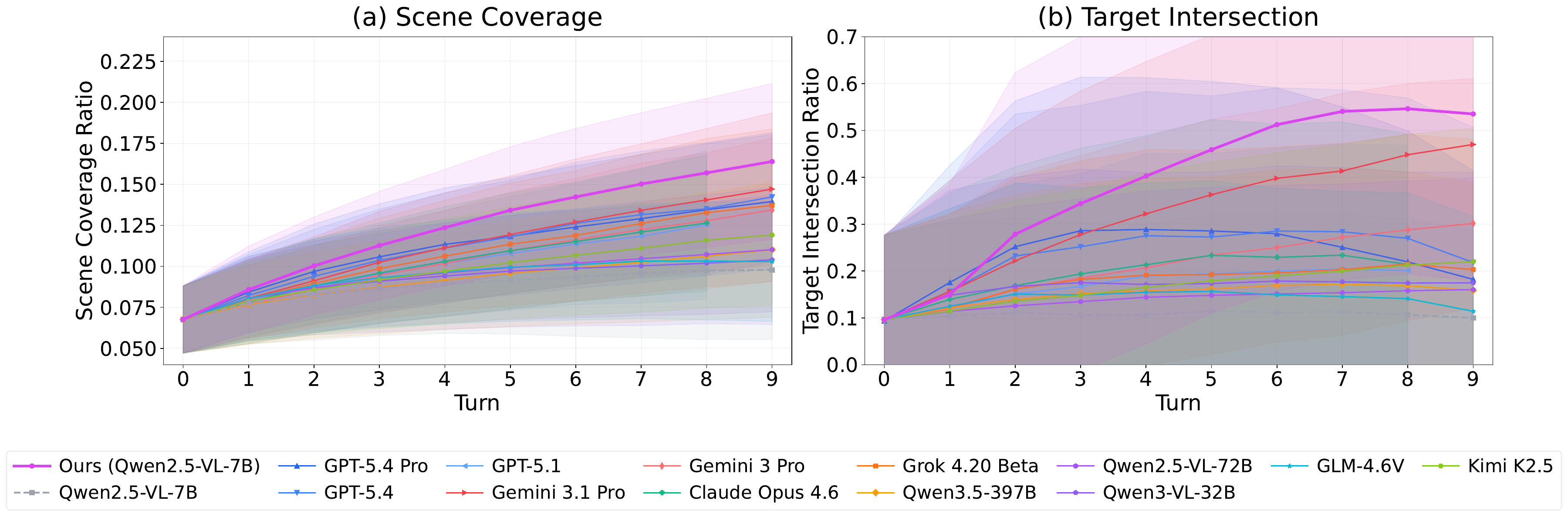}
\caption{Point cloud coverage on IVP across all $15$ models. (a)~Scene coverage ratio (fraction of all scene vertices observed). (b)~Target intersection ratio (fraction of target view vertices covered). Only our trained model sustains monotonic growth in target intersection ratio; frontier models plateau or decline after turns $5$--$7$.}
\label{fig:coverage_analysis_all}
\end{figure*}

Figure~\ref{fig:coverage_analysis_all} extends the coverage analysis from Section~\ref{sec:behavior} to all $15$ evaluated models. The pattern is consistent: our trained model is the only model that achieves sustained, monotonic growth in target intersection ratio across turns. Frontier proprietary models (GPT-5.4 Pro, Gemini 3.1 Pro, Gemini 3 Pro) show moderate initial increases but plateau or decline after turns $5$--$7$, suggesting they explore broadly without maintaining target-directed trajectories. Open-weight models (Qwen3.5-397B, Qwen2.5-VL-72B, Qwen3-VL-32B) generally remain below the proprietary models in both metrics.

\paragraph{Methodology.}
For each model, we collect $530$ rollout trajectories on the \viewsuite{} test set. At each turn, we render the agent's viewpoint against the scene's 3D point cloud and compute the set of visible vertices using depth-buffered rendering. We track two metrics:
\begin{itemize}[leftmargin=*,nosep]
\item \textbf{Scene coverage ratio}: $|\bigcup_{t=0}^{T} V_t| \,/\, |V_{\text{total}}|$, where $V_t$ is the set of vertices visible at turn $t$ and $V_{\text{total}}$ is the full scene point cloud.
\item \textbf{Target intersection ratio}: $|\bigcup_{t=0}^{T} V_t \cap V_{\text{target}}| \,/\, |V_{\text{target}}|$, where $V_{\text{target}}$ is the set of vertices visible from the ground-truth target viewpoint.
\end{itemize}
All models are evaluated for up to $10$ turns (turn~0 is the initial view). Some models produce fewer turns due to early stopping; turns with fewer than $1\%$ of the maximum trajectory count are excluded. Lines show per-turn means; shaded regions indicate $\pm 1$ standard deviation.

\begin{figure*}[h]
\centering
\includegraphics[width=\textwidth]{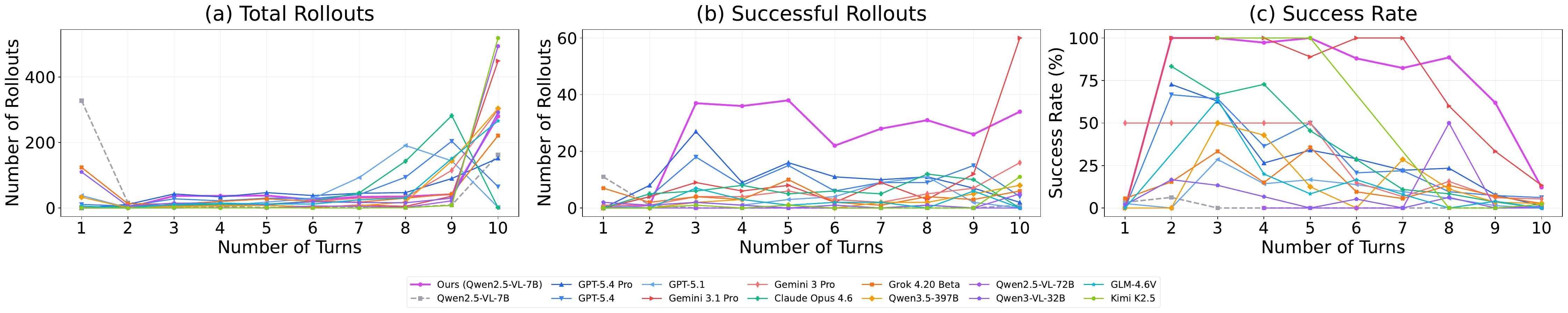}
\caption{Turn usage on IVP. (a)~Rollouts by turn count: the base model and GPT-5.4 Pro stop after a single turn on most rollouts, while our trained model and Gemini 3.1 Pro use the full $10$-turn budget. (b)~Successful rollouts and (c)~success rate by turn count: our trained model succeeds at high rates on short episodes.}
\label{fig:turn_distribution}
\end{figure*}

\paragraph{Turn distribution and success rate.}
Figure~\ref{fig:turn_distribution}(a) shows the distribution of turns used per rollout. The base Qwen2.5-VL-7B-Instruct and GPT-5.4 Pro terminate most rollouts after a single turn (no planning action taken). Our trained model and Gemini 3.1 Pro concentrate at $10$ turns (using all available turns).
Figure~\ref{fig:turn_distribution}(b,c) show successful rollouts and success rate by turn count. Our trained model achieves high success rates at low turn counts (efficient view planning on easy cases) with decreasing rates at higher turn counts (harder episodes).

\subsection{Attention Analysis: Methodology and Full Results}
\label{app:attention_method}

We measure the \textbf{image attention fraction} (fraction of response-token attention directed toward image tokens) across all $28$ layers on the same $530$ trajectories for both models.
Figure~\ref{fig:attention_all_layers} shows a consistent pattern: the trained model attends more to image tokens than the base model at nearly every layer, and this gap grows with depth (on average $1.4\times$ over layers $0$ to $8$, $1.9\times$ over layers $9$ to $18$, and $3.0\times$ over layers $19$ to $27$).
In other words, compared with the base model, the trained model maintains consistently higher attention to the image tokens throughout the layers, suggesting that it learns to use views as evidence for planning rather than relying mainly on textual or prior heuristics.

\paragraph{Setup.}
We run both our trained model and the base Qwen2.5-VL-7B on the same 530 trajectories.
For each trajectory, we perform a single forward pass and extract head-averaged attention from every layer.
We identify image token positions via the processor's token-to-image mapping and compute the \textbf{image attention fraction}: for each response token at position $q$, we measure $\sum_{k \in \mathcal{I}} \alpha_{q,k} \,/\, \sum_{k} \alpha_{q,k}$, where $\mathcal{I}$ is the set of image token indices and $\alpha_{q,k}$ is the attention weight.
We report this fraction averaged across response tokens within each turn.
Trajectories with fewer than $3$ images or no response turns are excluded.
Error bars indicate $\pm 1$ standard deviation across trajectories.

\paragraph{Implementation.}
Qwen2.5-VL-7B has $28$ transformer layers. We set \emph{all} layers to ``eager'' attention so that every layer uses the same computation. Extracting all $28$ layers at once exceeds single-GPU memory on long trajectories (up to roughly $6$K tokens and $12$ images), so we compute attention in tiles over the query dimension: the computation is exactly eager (a softmax over scaled $QK^\top$ scores), but only one query tile is materialized at a time, and we reduce each tile to the per-token image attention fraction on the fly.
All $28$ layers are therefore read from a single, fully consistent forward pass per trajectory, and both models use the identical procedure.
The full per-layer results are shown in Figure~\ref{fig:attention_all_layers} in the main text.

\subsection{Spatial Prior Transfer: Post-Training Details}
\label{app:downstream_transfer}

For spatial prior transfer experiments (Section~\ref{sec:transfer}), we post-train both our trained model and the base Qwen2.5-VL-7B-Instruct using GRPO with identical hyperparameters on each task.
We evaluate on three tasks:
\begin{itemize}[leftmargin=*,nosep]
\item \textbf{P2V} and \textbf{V2P}: view-action understanding tasks from \viewsuite{}, requiring understanding of how the viewpoint changes under actions. These are trained jointly from the same data splits.
\item \textbf{MindCube}~\citep{mindcube}: mental rotation and spatial simulation, requiring the model to track object transformations across viewpoints.
\end{itemize}

\paragraph{Training hyperparameters.}
All tasks use GRPO with $n{=}8$ rollouts per prompt, actor learning rate $1 \times 10^{-6}$, and KL penalty disabled ($\lambda_{\text{KL}} = 0$).
Training runs for $401$ steps on $8$ GPUs with FSDP and gradient checkpointing.
Table~\ref{tab:downstream_hyperparams} summarizes per-task differences.

\begin{table}[h]
\centering
\caption{Per-task hyperparameters for downstream transfer post-training. All other settings are shared (see text).}
\label{tab:downstream_hyperparams}
\small
\setlength{\tabcolsep}{4pt}
\begin{tabular}{@{}lccc@{}}
\toprule
& P2V / V2P & MindCube \\
\midrule
Train batch size & 64 & 32 \\
Max prompt length & 4{,}000 & 3{,}000 \\
Max response length & 2{,}000 & 4{,}000 \\
\bottomrule
\end{tabular}
\end{table}

\paragraph{Reward functions.}
All tasks use a binary reward composed of a format reward and an answer reward.
The model must produce a valid response in the format \texttt{...<action>...</action>}.
For P2V, V2P, and MindCube, the answer reward checks whether the predicted option letter matches the ground truth (case-insensitive first-character match).
The reward weights are:
\begin{itemize}[leftmargin=*,nosep]
\item P2V / V2P: $r = 0.1 \cdot r_{\text{format}} + 0.9 \cdot r_{\text{answer}}$
\item MindCube: $r = 0.2 \cdot r_{\text{format}} + 0.8 \cdot r_{\text{answer}}$
\end{itemize}

Both models are trained for the same number of steps on the same data to ensure a fair comparison of the spatial priors each model brings.

\end{document}